\documentclass{article}
\usepackage[square,numbers]{natbib}

\usepackage[preprint]{neurips_2026}
\usepackage{amsmath}
\usepackage{amssymb}
\newcommand{\R}{\mathbb{R}} 

\usepackage[utf8]{inputenc} 
\usepackage[T1]{fontenc}    
\usepackage{hyperref}       
\usepackage{url}            
\usepackage{booktabs}       
\usepackage{amsfonts}       
\usepackage{nicefrac}       
\usepackage{microtype}      
\usepackage{xcolor}         
\usepackage{algorithm}
\usepackage{algpseudocode}

\usepackage{graphicx}
\usepackage{grffile}
\usepackage{subcaption}
\usepackage{pifont}
\usepackage{booktabs}
\usepackage{tikz}
\usepackage{amsmath,amssymb}
\usepackage{tikz}
\usetikzlibrary{positioning,arrows.meta,shapes.geometric,calc,matrix,backgrounds,fit}
\usetikzlibrary{matrix, positioning, arrows.meta, backgrounds, calc}
\usepackage{adjustbox}
\usepackage{pgfplots}
\pgfplotsset{compat=1.18}
\usepackage[capitalize,noabbrev]{cleveref}
\usepackage[textsize=tiny]{todonotes}
\usepackage{multirow}
\usepackage{amsmath,amssymb,mathtools,bm,empheq,amsthm}
\usepackage{arydshln}
\usepackage{pifont}
\usepackage{pgfplots}
\usepackage{makecell}
\pgfplotsset{compat=1.17}
\usetikzlibrary{pgfplots.groupplots}
\usepgfplotslibrary{groupplots}
\usepackage{tikz}
\usetikzlibrary{plotmarks}
\usepackage{xcolor}         
\title{GeoPair: Geometry-Preserving Cross-Layer Factorization for Training-Free Transformer Compression}

\theoremstyle{plain}

\theoremstyle{definition}

\theoremstyle{remark}

\def\bal#1\eal{\begin{align}#1\end{align}} 

\newcommand{\pr}[1]{\left(#1\right)} 
\DeclareMathOperator*{\argmin}{arg\,min} 
\DeclareMathOperator*{\argmax}{arg\,max} 

\def\transp{\mathsf{T}} 
\def\m{\mathbf}

\def\R{\mathbb{R}}

\newcommand{\norm}[2]{\ensuremath{\left\|#1\right\|_{#2}}}

\newcommand {\bbmtx}{\begin{bmatrix}} 
\newcommand {\ebmtx}{\end{bmatrix}} 

\DeclareMathOperator*{\diagonal}{diag} 
\newcommand{\diag}[1]{\diagonal\pr{#1}}

\definecolor{my_red}{rgb}{0.98, 0.20, 0.14}  
\definecolor{my_green}{rgb}{0.15,0.95,0.15}  %
\definecolor{my_white}{rgb}{1,1,1}  %

\newcommand{\wh}[1]{\colorbox{my_white}{#1}}
\newcommand{\ours}{GeoPair}
\newcommand{\cmark}{\ding{51}}%
\newcommand{\xmark}{\ding{55}}%
\author{%
  \textbf{Baher Mohammad$^{1,2*}$,}
  \textbf{Ammar Ali$^{1,2*}$,}
  \textbf{Stamatios Lefkimmiatis$^{1,2}$} \\
  $^{1}$MWS AI $\quad$ $^{2}$ITMO \\
  \texttt{baher@itmo.ru, ammarali32@itmo.ru, s.lefkimmiatis@mts.ai}
}

\begin{document}

\maketitle

\begingroup
\renewcommand{\thefootnote}{\fnsymbol{footnote}}
\footnotetext[1]{Equal contribution.}
\endgroup

\begin{abstract}

Transformer architectures exhibit cross-layer redundancies, yet post-training compression pipelines typically optimize layers in isolation or rely on heuristic grouping strategies that disregard layer-specific activation geometries. We introduce a principled, training-free framework that sequentially optimizes cross-layer weight pairings and shared-dictionary factorizations. Rather than forcing weights of adjacent layers to share a basis or heuristically merging activation statistics, our approach identifies structurally compatible projections and learns a shared representation that better preserves each layer’s distinct calibration geometry. Coupled with structured sparsity, this yields highly efficient weight decompositions without sacrificing functional fidelity. Across diverse architectures, scales, and modalities, our method achieves state-of-the-art results, consistently outperforming independent structured weight decompositions and alternative pairwise weight factorizations, which operate under heuristic grouping strategies. By combining geometry-preserving dictionary updates
with graph-based pairing and sparse coefficient
optimization, we provide a principled framework
for transformer compression across modalities.
\end{abstract}

\section{Introduction}
The widespread adoption of transformer-based architectures has yielded unprecedented capabilities across language \cite{llama, qwen, gpt, gemma, phi}, vision \cite{detr, vit}, and generative tasks \cite{wan2025wan0}. However, their substantial memory footprint and computational overhead present a critical bottleneck for deployment in resource-constrained environments. Post-training model compression has emerged as a practical alternative to retraining, with matrix factorization methods offering a compelling trade-off between parameter efficiency and functional fidelity. While conventional approaches approximate each layer's weight matrix independently, they overlook a fundamental structural property: substantial cross-layer redundancies emerge naturally across deep transformer stacks.

Exploiting such redundancies through shared-dictionary learning, (where multiple projections are represented by a common dictionary with layer-specific coefficients), promises sublinear storage scaling without sacrificing expressiveness. Yet, principled pairwise cross-layer sharing remains largely unexplored in post-training compression. The primary obstacle lies in the data-aware nature of modern factorization pipelines: accurate low-rank approximation requires a whitening transform calibrated to each layer's activation distribution. Because these distributions vary significantly across depths, their associated whitening spaces are incompatible, complicating direct dictionary sharing. Recent approaches such as \cite{basis_sharing} heuristically aggregate layer-wise covariances into global whitening transforms and restrict sharing to adjacent layers. This introduces two limitations: (i) covariance averaging distorts layer-specific activation geometry, degrading fidelity; (ii) fixed adjacency ignores non-local alignments, leaving gains unrealized.

In this work, we introduce a principled, training-free framework that optimizes cross-layer grouping and shared-dictionary factorization under layer-specific whitening transforms via alternating minimization. Rather than relying on heuristic covariance merging or greedy adjacency rules, our method identifies structurally compatible projections and learns a shared representation that rigorously preserves each layer's calibration geometry. We formulate the dictionary update as a generalized Sylvester equation, enabling closed-form solutions that respect distinct whitening spaces. Layer pairing is cast as a maximum-weight matching problem, solved via Edmonds' Blossom algorithm \cite{blossom} using a shape-agnostic column-space alignment metric, with the solution being optimal with respect to the proposed distance metric. To further enhance compression efficiency and adapt the method for dictionary learning-based decompositions, we utilize Hard Thresholding Pursuit (HTP) \cite{htp} powered by a conjugate gradient linear solver, enabling structured coefficient sparsity without heuristic budget allocation or dynamic scheduling.

\textbf{Contributions}: Our framework establishes a reproducible, optimization-driven alternative to heuristic compression pipelines. The main contributions are:
\begin{itemize}
\item \textbf{Shared-dictionary learning under distinct whitening spaces}: We introduce a closed-form generalized Sylvester solver that eliminates heuristic covariance aggregation while preserving layer-specific activation geometry.
\item \textbf{Layer pairing:} We formulate cross-layer grouping as a weighted maximum matching problem, replacing fixed adjacency heuristics with a data-driven strategy that minimizes structural discrepancy across the entire architecture.
\item \textbf{Sparse coefficient optimization:} We extend the framework to sparse dictionary learning via HTP, enabling adaptive, layer-specific compression that outperforms dense low-rank baselines at high compression ratios.
\item \textbf{Broad empirical validation:} Extensive experiments across diverse architectures, scales, and modalities demonstrate state-of-the-art results, consistently outperforming independent factorization, heuristic merging, and existing dictionary learning approaches.
\end{itemize}

By unifying Layer grouping, closed-form dictionary updates, and sparse coding, our work provides a theoretically grounded foundation for scalable, transformers compression.\vspace{-.3cm}

\section{Related Work}\vspace{-.2cm}
\paragraph{Data-Aware Matrix Factorization.}
Post-training compression via matrix factorization has emerged as a practical strategy for reducing transformer memory footprint without fine-tuning. Truncated singular value decomposition (SVD) yields the optimal rank-$r$ approximation of a weight matrix under the Frobenius norm, and is mathematically equivalent to performing principal component analysis (PCA) on its column space \cite{bishop2006pattern}. Early compression pipelines applied this decomposition directly to pretrained weights, but assumed isotropic activation statistics, ignoring the input-dependent scaling inherent to transformer forward passes. This geometric mismatch causes significant accuracy degradation at high compression ratios. Subsequent data-aware approaches~\cite{drone, asvd, svd_llm, masa} aligned the factorization objective with true activation reconstruction by operating in a calibration-induced whitened space. While these methods preserve downstream performance, they treat each layer in isolation, overlooking the substantial cross-layer redundancy inherent in deep architectures.

\paragraph{Dictionary Learning and Sparse Decomposition.}
To improve compression fidelity, recent work has shifted from dense low-rank approximations to dictionary learning formulations, which decouple a shared dictionary from layer-specific coefficient matrices. Classical sparse coding algorithms such as $k$-SVD~\cite{k_svd} and Method of Optimal Directions~\cite{mod} have been adapted to the transformer compression setting. Methods like CoSpaDI~\cite{cospadi}, ROCKET \cite{rocket}, and COMPOT~\cite{compot} demonstrate that structured sparsity often yields superior trade-offs compared to dense baselines, particularly when coefficient matrices are heavily constrained. However, these approaches still optimize coefficients and dictionaries per layer or rely on heuristic update schedules, leaving explicit cross-layer parameter sharing unexplored.

\paragraph{Cross-Layer Grouping and Shared Basis Learning.}
Explicitly grouping structurally similar layers to learn a shared basis offers a promising path to sublinear storage scaling. The most notable advance in this direction, Basis Sharing~\cite{basis_sharing} pairs adjacent layers and learns a joint basis, outperforming independent baselines. Yet, this approach suffers from three critical limitations: (i) fixed adjacency for grouping ignoring non-local alignments; (ii) global covariance aggregation which distorts layer-specific geometry; (iii) manual, model-specific constraints limit universal applicability. (e.g., restricting sharing between certain projection types), meaning the approach cannot be applied universally to maintain numerical stability. Complementary work like Matrix PCA~\cite{masa} extracts shared bases via Eigen Value Decomposition (EVD) on stacked weights but remains constrained by distributional-drift-based grouping and independent refinement stages, necessitating manual budgeting and failing to optimize grouping under distinct whitening geometries.

\paragraph{Positioning of Our Approach.}
Our framework addresses these gaps through a principled, optimization-driven pipeline that sequentially solves layer grouping and shared-dictionary learning. Rather than heuristic covariance merging, we formulate cross-layer dictionary sharing under distinct whitening transforms as a generalized Sylvester equation, yielding closed-form dictionary updates that preserve individual layer geometries. We replace fixed adjacency rules with a global maximum-weight matching strategy~\cite{blossom}, optimally pairing layers with respect to a shape-agnostic column-space alignment metric. Finally, we integrate Hard Thresholding Pursuit (HTP)~\cite{htp} with conjugate gradient  to enforce structured coefficient sparsity, enabling flexible compression without heuristic budget allocation or dynamic scheduling. This eliminates manual engineering while establishing a theoretically grounded pathway for scalable, multi-modal model compression.\vspace{-.3cm}

\section{Method}
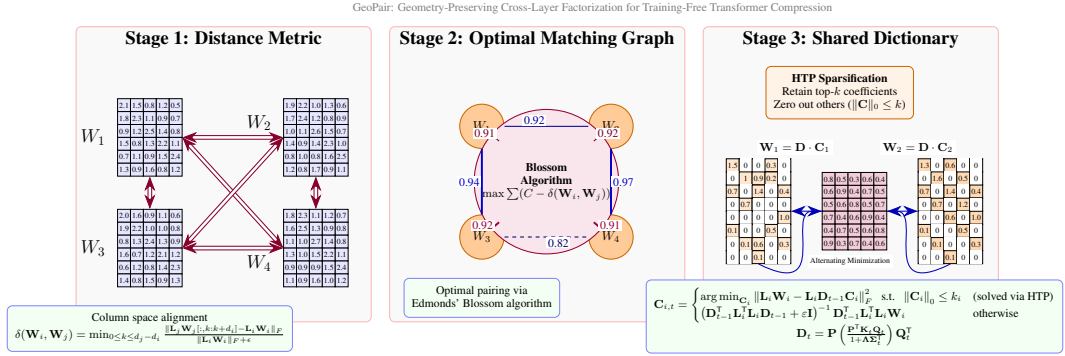
\begin{figure}[htbp]
\vspace{-.5cm}
\centering

\begin{adjustbox}{width=\textwidth,center}

\begin{tikzpicture}[
    matrixnode/.style={
        matrix of nodes,
        nodes={draw, fill=blue!10, minimum size=0.35cm, anchor=center, font=\tiny},
        column sep=-\pgflinewidth,
        row sep=-\pgflinewidth,
        inner sep=0pt,
        outer sep=0pt
    },
    dictnode/.style={
        matrix of nodes,
        nodes={draw, fill=purple!20, minimum size=0.35cm, anchor=center, font=\tiny},
        column sep=-\pgflinewidth,
        row sep=-\pgflinewidth,
        inner sep=0pt,
        outer sep=0pt
    },
    coefnode/.style={
        matrix of nodes,
        nodes={draw, fill=orange!20, minimum size=0.35cm, anchor=center, font=\tiny},
        column sep=-\pgflinewidth,
        row sep=-\pgflinewidth,
        inner sep=0pt,
        outer sep=0pt
    },
    sparsecoef/.style={
        matrix of nodes,
        nodes={draw, fill=orange!30, minimum size=0.35cm, anchor=center, font=\tiny},
        column sep=-\pgflinewidth,
        row sep=-\pgflinewidth,
        inner sep=0pt,
        outer sep=0pt
    },
    graphnode/.style={
        circle, 
        draw=orange!80!black, 
        fill=orange!40, 
        minimum size=1.2cm,
        thick,
        font=\bfseries
    },
    centralnode/.style={
        circle, 
        draw=purple!80!black, 
        fill=purple!10, 
        minimum size=2cm,
        thick,
        align=center,
        font=\bfseries\small
    },
    arrowstyle/.style={
        -{Stealth[length=4mm,width=3mm]},
        thick,
        blue!70!black
    },
    doublearrow/.style={
        {Stealth[length=4mm,width=3mm]}-{Stealth[length=4mm,width=3mm]},
        thick,
        purple!70!black,
        double,
        double distance=2pt
    },
    stagebox/.style={
        rectangle,
        draw=red!30,
        fill=gray!5,
        rounded corners,
        minimum width=8cm,
        minimum height=8cm,
        align=center,
        font=\bfseries\Large
    },
    info box/.style={
        draw=blue!50,
        fill=green!5,
        rounded corners,
        inner sep=6pt,
        align=center,
        font=\small
    },
    htpbox/.style={
        draw=orange!80!black,
        fill=orange!10,
        rounded corners,
        inner sep=6pt,
        align=center,
        font=\small
    }]

\node[font=\bfseries\Huge, text=blue!80!black] (maintitle) at (8.5,9.5) 
    {};
\node[above=0.1cm of maintitle, font=\small, text=gray] 
    {GeoPair: Geometry-Preserving Cross-Layer Factorization for Training-Free Transformer Compression};

\node[stagebox] (stage1box) at (-1.5,5.5) {};
\node[font=\bfseries\Large, anchor=north] (stage1title) at (stage1box.north) {Stage 1: Distance Metric};

\matrix[matrixnode] (W1) at (-3.5,6.5) {
    2.1 & 1.5 & 0.8 & 1.2 & 0.5 \\
    1.8 & 2.3 & 1.1 & 0.9 & 0.7 \\
    0.9 & 1.2 & 2.5 & 1.4 & 0.8 \\
    1.5 & 0.8 & 1.3 & 2.2 & 1.1 \\
    0.7 & 1.1 & 0.9 & 1.5 & 2.4 \\
    1.3 & 0.9 & 1.6 & 0.8 & 1.2 \\
};
\node[left=0.2cm of W1, font=\Large\bfseries] {$W_1$};

\matrix[matrixnode] (W2) at (1.0,6.5) {
    1.9 & 2.2 & 1.0 & 1.3 & 0.6 \\
    1.7 & 2.4 & 1.2 & 0.8 & 0.9 \\
    1.0 & 1.1 & 2.6 & 1.5 & 0.7 \\
    1.4 & 0.9 & 1.4 & 2.3 & 1.0 \\
    0.8 & 1.0 & 0.8 & 1.6 & 2.5 \\
    1.2 & 0.8 & 1.7 & 0.9 & 1.1 \\
};
\node[left=0.2cm of W2, yshift=0.4cm, font=\Large\bfseries] {$W_2$};

\matrix[matrixnode] (W3) at (-3.5,3.5) {
    2.0 & 1.6 & 0.9 & 1.1 & 0.6 \\
    1.9 & 2.2 & 1.0 & 1.0 & 0.8 \\
    0.8 & 1.3 & 2.4 & 1.3 & 0.9 \\
    1.6 & 0.7 & 1.2 & 2.1 & 1.2 \\
    0.6 & 1.2 & 0.8 & 1.4 & 2.3 \\
    1.4 & 0.8 & 1.5 & 0.9 & 1.3 \\
};
\node[left=0.2cm of W3, font=\Large\bfseries] {$W_3$};

\matrix[matrixnode] (W4) at (1.0,3.5) {
    1.8 & 2.3 & 1.1 & 1.2 & 0.7 \\
    1.6 & 2.5 & 1.3 & 0.9 & 0.8 \\
    1.1 & 1.0 & 2.7 & 1.4 & 0.8 \\
    1.3 & 1.0 & 1.5 & 2.2 & 1.1 \\
    0.9 & 0.9 & 0.9 & 1.5 & 2.4 \\
    1.1 & 0.9 & 1.6 & 1.0 & 1.2 \\
};
\node[left=0.2cm of W4, yshift=-0.4cm, font=\Large\bfseries] {$W_4$};


\draw[doublearrow] (W1.south) -- (W3.north);
\draw[doublearrow] (W1.east) -- (W2.west);
\draw[doublearrow] (W3.east) -- (W4.west);
\draw[doublearrow] (W2.south) -- (W4.north);
\draw[doublearrow] (W1.east) -- (W4.west);
\draw[doublearrow] (W3.east) -- (W2.west);
\node[info box, below=0.5cm of W3.south ] {Column space alignment\\ $
\delta(\m W_i, \m W_j) = \min_{0 \le k \le d_j - d_i} \frac{\| \m L_j \m W_j[:, k:k+d_i] - \m L_i \m W_i \|_F}{\| \m L_i \m W_i \|_F + \epsilon}
$};


\node[stagebox] (stage2box) at (7,5.5) {};
\node[font=\bfseries\Large, anchor=north] (stage2title) at (stage2box.north) {Stage 2: Optimal Matching Graph};

\node[graphnode] (G1) at (5.5,6.8) {$W_1$};
\node[graphnode] (G2) at (9,6.8) {$W_2$};
\node[graphnode] (G3) at (5.5,3.8) {$W_3$};
\node[graphnode] (G4) at (9,3.8) {$W_4$};

\node[centralnode] (blossom) at (7.25,5.3) {Blossom\\Algorithm\\$\max \sum (C - \delta(\m W_i, \m W_j))$};

\draw[thick, blue!60!black] (G1) -- node[above left, font=\small, fill=white, inner sep=1pt] {0.92} (G2);
\draw[thick, blue!80!black, line width=1.5pt] (G1) -- node[left, font=\small, fill=white, inner sep=1pt] {0.94} (G3);
\draw[thick, blue!80!black, line width=1.5pt] (G2) -- node[right, font=\small, fill=white, inner sep=1pt] {0.97} (G4);
\draw[thick, blue!40!black, dashed] (G3) -- node[below right, font=\small, fill=white, inner sep=1pt] {0.82} (G4);
\draw[thick, purple!60!black, dashed] (G1) -- node[above left, font=\small, pos=0.3, fill=white, inner sep=1pt] {0.91} (blossom);
\draw[thick, purple!60!black, dashed] (G2) -- node[above right, font=\small, pos=0.3, fill=white, inner sep=1pt] {0.92} (blossom);
\draw[thick, purple!80!black, line width=1.5pt] (G3) -- node[left, font=\small, pos=0.5, fill=white, inner sep=1pt] {0.92} (blossom);
\draw[thick, purple!80!black, line width=1.5pt] (G4) -- node[right, font=\small, pos=0.5, fill=white, inner sep=1pt] {0.91} (blossom);

\begin{scope}[on background layer]
    \fill[blue!15, opacity=0.5] (G1.center) -- (G3.center) -- (blossom.center) -- cycle;
    \fill[blue!15, opacity=0.5] (G2.center) -- (G4.center) -- (blossom.center) -- cycle;
\end{scope}

\node[info box, below=0.5cm of G3.south] {Optimal pairing via\\Edmonds' Blossom algorithm};


\node[stagebox] (stage3box) at (15.5,5.5) {};
\node[font=\bfseries\Large, anchor=north] (stage3title) at (stage3box.north) {Stage 3: Shared Dictionary};

\matrix[coefnode] (C1) at (13,4.5) {
1.5 & \wh{0} & \wh{0} & 0.3 & \wh{0} \\
\wh{0} & 1 & 0.9 & 0.2 & \wh{0} \\
0.7 & \wh{0} & 1.4 & \wh{0} & 0.4 \\
\wh{0} & 0.7 & \wh{0} & \wh{0} & \wh{0} \\
\wh{0} & \wh{0} & \wh{0} & \wh{0} & 1.0 \\
0.1 & \wh{0} & \wh{0} & 0.5 & \wh{0} \\
\wh{0} & 0.1 & 0.6 & \wh{0} & 0.3 \\
\wh{0} & \wh{0} & 0.1 & \wh{0} & \wh{0} \\
};

\matrix[dictnode] (D) at (15.6,4.5) {
    0.8 & 0.5 & 0.3 & 0.6 & 0.4 \\
    0.6 & 0.9 & 0.4 & 0.7 & 0.5 \\
    0.5 & 0.6 & 0.8 & 0.5 & 0.7 \\
    0.7 & 0.4 & 0.6 & 0.9 & 0.4 \\
    0.4 & 0.7 & 0.5 & 0.6 & 0.8 \\
    0.9 & 0.3 & 0.7 & 0.4 & 0.6 \\
};

\matrix[sparsecoef] (C2) at (18.2,4.5) {
1.3 & \wh{0} & 0.6 & \wh{0} & \wh{0} \\
\wh{0} & 1.6 & \wh{0} & 0.5 & \wh{0} \\
0.7 & \wh{0} & 1.4 & \wh{0} & 0.4 \\
\wh{0} & 0.7 & \wh{0} & 1.2 & \wh{0} \\
\wh{0} & \wh{0} & 0.6 & \wh{0} & 1.0 \\
0.1 & \wh{0} & \wh{0} & 0.5 & \wh{0} \\
\wh{0} & 0.1 & \wh{0} & \wh{0} & 0.3 \\
\wh{0} & \wh{0} & 0.1 & \wh{0} & \wh{0} \\
};

\node[font=\small, fill=white, inner sep=2pt] at ($(C1.north east)+(0.1,0.3)$) {$\m W_1 = \m D \cdot \m C_1$};

\node[font=\small, fill=white, inner sep=2pt] at ($(C2.north west)+(0.0,0.3)$) {$\m W_2 = \m D \cdot \m C_2$};

\node[font=\tiny, fill=white, inner sep=2pt] at ($(D.south east)+(-1.0,-0.3)$) {Alternating Minimization};

\node[htpbox, below=-2.5cm of C2.north, xshift=-3.0cm] (htpbox) {
    \textbf{HTP Sparsification}\\
    Retain top-$k$ coefficients\\
    Zero out others ($\|\m C\|_0 \leq k$)
};

\draw[arrowstyle, bend left=15] (D.west) -- (C1.east);
\draw[arrowstyle] (C1.south) .. controls +(2, -1) and +(-1, 0) .. (D.west);
\draw[arrowstyle, bend right=15] (D.east) -- (C2.west);
\draw[arrowstyle] (C2.south) .. controls +(-2, -1) and +(1, 0) .. (D.east);
\node[info box, below=0.8cm of D.south] {
$\m C_{i,t} = \begin{cases}
       \argmin_{\m C_i} \norm{\m L_i \m W_i - \m L_i \m D_{t-1} \m C_i}{F}^2 \;\; \text{s.t.} \;\; \norm{\m C_i}{0} \leq k_i & \text{(solved via HTP)} \\
       \pr{\m D_{t-1}^\transp \m L_i^\transp \m L_i \m D_{t-1} + \varepsilon \m I}^{-1} \m D_{t-1}^\transp \m L_i^\transp \m L_i \m W_i & \text{otherwise}
       \end{cases}$ \\
       $\m D_{t} = \m P \left( \frac{\m P^\transp \m K_t \m Q_t}{1 + \bm\Lambda \bm\Sigma_t^\transp} \right) \m Q_t^\transp$
};

\end{tikzpicture}

\end{adjustbox}

\caption{\textbf{GeoPair framework overview.} Stage 1 computes whitened-space structural distances between 
candidate weight matrices. Stage 2 solves for optimal pairings via maximum-weight graph matching with respect to the distance metric. 
Stage 3 learns shared dictionaries for each pair through calibration-aware alternating minimization with 
optional HTP-based coefficient sparsification.}
\label{fig:framework}
\vspace{-0.2 cm}
\end{figure}
\subsection{Overview and Problem Setup}
\label{sec:problem_setup}
Similar to prior post-training compression work~\cite{svd_llm}, we formulate compression as \emph{activation reconstruction} over a small calibration set. We consider a pretrained transformer with linear projections parameterized by weight matrices $\m W \in \R^{d \times d_{out}}$ and seek a structured approximation $\widehat{\m W}$ that reduces storage and computation while preserving functional behavior, without the need for finetuning using back-propagation.

Let $\m X \in \R^{N \times d}$ denote calibration activations and define the empirical Gram matrix $\m G = \m X^\transp \m X$. In practice, limited calibration data often yields a rank-deficient $\m G$, rendering it singular. To guarantee a well-posed whitening transform, we enforce non-singularity by introducing a Tikhonov regularizer $\m G_\eta = \m X^\transp \m X + \eta \m I$ ($\eta > 0$). This is mathematically equivalent to augmenting the reconstruction objective with a pure weight-space $\ell_2$ penalty, which strictly ensures $\m G_\eta \succ 0$ and admits a unique Cholesky factorization as a whitening transformation $\m G_\eta = \m L^\transp \m L$. The activation reconstruction objective is then equivalently written as
\bal
\widehat{\m W}
=
\argmin_{\widehat{\m W}}
\norm{\m X \pr{\m W - \widehat{\m W}}}{F}^2 + \eta \norm{\m W - \widehat{\m W}}{F}^2
=
\argmin_{\widehat{\m W}}
\norm{\m L \pr{\m W - \widehat{\m W}}}{F}^2 .
\eal
This shows that minimizing the activation reconstruction error is equivalent to minimizing the reconstruction in the whitened space induced by calibration statistics.
\subsection{Cross-Layer Shared Dictionary Optimization}
\label{sec:paired_objective}

Unlike standard compression pipelines that optimize each projection in isolation, we aim to exploit structural redundancies
that naturally arise across layers sharing the same input dimension. By coupling their factorizations, we can learn a single
shared dictionary that efficiently spans both layers, yielding higher compression ratios at comparable reconstruction fidelity.

Building on this motivation, our optimization objective remains strictly tied to minimizing functional activation error.
Following the equivalence established in Section~\ref{sec:problem_setup}, preserving the input--output behavior for two
compatible projections, translates directly to minimizing their respective calibration-weighted reconstruction errors.\vspace{-.3cm}

\paragraph{Cross-Layer Shared Dictionary Formulation.}
Consider two weight matrices $\m W_1 \in \R^{d \times d_1}$ and $\m W_2 \in \R^{d \times d_2}$, each with its own
layer-specific Cholesky whitening transform $\m L_1, \m L_2 \in \R^{d \times d}$. We approximate them using a
\emph{shared dictionary} $\m D \in \R^{d \times r}$ (where $r \ll d$) and layer-specific coefficient matrices
$\m C_1 \in \R^{r \times d_1}$, $\m C_2 \in \R^{r \times d_2}$.
We define:
\[
s_i=\norm{\m L_i\m W_i}{F},
\qquad
\bar{\m L}_i=\frac{\m L_i}{s_i},
\qquad
\bar{\m G}_i
=\bar{\m L}_i^{\transp}\bar{\m L}_i
=\frac{\m L_i^{\transp}\m L_i}{s_i^2}.
\]
We assume nonzero weight matrices, so $s_i>0$.
Our objective is to minimize the sum of the two squared
relative Frobenius reconstruction errors, which can be
written using the normalized whitening transforms as
\begin{align}
\min_{\m D,\m C_1,\m C_2}\quad
&\sum_{i=1}^{2}
\norm{\bar{\m L}_i(\m W_i-\m D\m C_i)}{F}^{2}.
\label{eq:joint_objective}
\end{align}
The shared dictionary $\m D$ captures common directional components activated across both layers, while $\m C_i$ projects these
components onto each layer's output space. After compression, the original parameter space is recovered via
$\widehat{\m W}_i= \m D \m C_i$.\vspace{-.2cm}

\paragraph{Alternating Minimization.}
The objective in Eq.~\eqref{eq:joint_objective} is bi-convex in $\pr{\m D, \m C}$. We optimize it via alternating minimization, which decouples the joint problem into two sequential subproblems. Because the optimization alternates between the dictionary and the coefficients, we require a distinct closed-form update rule for each block. Hereafter, we first derive the update rule for the coefficients $\m C_1, \m C_2$ given a fixed $\m D_{t-1}$, and then present the update rule for $\m D_t$ given the newly computed coefficients. These two steps are applied cyclically until convergence.

\textbf{Coefficient update
($\m C_1,\m C_2$ given $\m D_{t-1}$).}
With $\m D_{t-1}$ fixed, we compute the coefficients
using ridge-regularized least squares:
\begin{align}
\m C_{i,t}
&=
\argmin_{\m C_i}
\left\{
\norm{
\bar{\m L}_i(\m W_i-\m D_{t-1}\m C_i)
}{F}^{2}
+\varepsilon\norm{\m C_i}{F}^{2}
\right\}
\nonumber\\
&=
\left(
\m D_{t-1}^{\transp}\bar{\m G}_i\m D_{t-1}
+\varepsilon\m I
\right)^{-1}
\m D_{t-1}^{\transp}\bar{\m G}_i\m W_i,
\qquad i\in\{1,2\}.
\label{eq:coefficient_update}
\end{align}
Here $\varepsilon=10^{-6}$ regularizes the coefficient
subproblem and ensures a unique minimizer.
In practice, we solve the corresponding linear
system without explicitly forming the inverse.

\textbf{Dictionary update
($\m D_t$ given $\m C_{1,t},\m C_{2,t}$).}
With the coefficients fixed, we solve the regularized
dictionary subproblem
\begin{equation}
J_{\mathrm{rel}}(\m D)
=
\sum_{i=1}^{2}
\left[
\norm{
\bar{\m L}_i(\m W_i-\m D\m C_{i,t})
}{F}^{2}
+
\epsilon\norm{\bar{\m L}_i\m D}{F}^{2}
\right].
\label{eq:dictionary_relative_objective}
\end{equation}
Setting the derivative of $J_{\mathrm{rel}}$ to zero
gives the generalized Sylvester normal equation
\begin{equation}
\bar{\m G}_1\m D_t\tilde{\m B}_{1,t}
+
\bar{\m G}_2\m D_t\tilde{\m B}_{2,t}
=
\m K_t,
\label{eq:dictionary_normal}
\end{equation}
where
\begin{equation}
\m K_t
=
\sum_{i=1}^{2}
\bar{\m G}_i\m W_i\m C_{i,t}^{\transp}.
\end{equation}
Defining $\Phi(\cdot, \cdot)$ as the \emph{generalized eigenvalue decomposition} operator, we compute the activation Gram pair decomposition \emph{once} to obtain constant transformation matrices $(\m P,\bm\Lambda)
=\Phi(\bar{\m G}_2,\bar{\m G}_1)$. At each iteration $t$, we stabilize the coefficient Gram matrices via $\tilde{\m B}_{i,t} = \m C_{i,t}\m C_{i,t}^\top + \epsilon \m I$ 
and decompose the regularized pair to obtain $(\m Q_t, \bm \Sigma_t) = \Phi(\tilde{\m B}_{2,t}, \tilde{\m B}_{1,t})$. Using these transformations we can decouple the Sylvester equation into independent scalar equations, yielding the closed-form solution:
\bal
\m D_t
=
\m P
\left(
\frac{\m P^\transp\m K_t\m Q_t}
{\mathbf{1}_d\mathbf{1}_r^\transp
+\bm\lambda\bm\sigma_t^\transp}
\right)
\m Q_t^\transp,
\label{d_upd}
\eal
where division is elementwise,
$\bm\lambda=\diag{\bm\Lambda}$,
$\bm\sigma_t=\diag{\bm\Sigma_t}$,
and $\mathbf{1}_d\in\R^d$ and
$\mathbf{1}_r\in\R^r$ are vectors of ones.
For more details we refer to
Appendix~\ref{full_theory}.
This simultaneous diagonalization approach provides a deterministic update for $\m D_t$ without iterative optimization or step-size tuning. 
The dense coefficient update exactly minimizes its
ridge-regularized quadratic subproblem.

\subsection{Sparse Matrix Coefficients via Hard Thresholding Pursuit}
\label{sec:htp_coefficients}

Recent state-of-the-art post-training compression methods increasingly rely on dictionary learning formulations that enforce structured sparsity in the factorized representations. Motivated by these advances, we integrate a sparsity-constrained coefficient update directly into our alternating minimization pipeline as a core mechanism for maximizing compression fidelity under strict parameter budgets.  Specifically, we replace the dense least-squares coefficient update step with an $\ell_0$-constrained formulation:
\begin{align}
\min_{\m C_{i,t}}\quad
&
\norm{
\bar{\m L}_i
(\m W_i-\m D_{t-1}\m C_{i,t})
}{F}^{2}
+\varepsilon\norm{\m C_{i,t}}{F}^{2}
\nonumber\\
\text{s.t.}\quad
&
\norm{\m C_{i,t}[:,j]}{0}\leq k,
\qquad j=1,\ldots,d_i.
\label{eq:sparse_coefficient_objective}
\end{align}
Here $k$ is the fixed per-column sparsity budget,
and $\norm{\cdot}{0}$ counts nonzero entries.
We approximately solve this subproblem using
Hard Thresholding Pursuit (HTP).
We define
$\mathrm{CR}=1-S_{\mathrm{comp}}/S_{\mathrm{raw}}$,
where $S_{\mathrm{raw}}$ and $S_{\mathrm{comp}}$
denote the nominal uncompressed and compressed
representation sizes, respectively.
The ratio $\mathrm{KS}=r/k$ is the inverse
nonzero fraction per coefficient column
(see Appendix~\ref{app:budget}).

For a fixed dictionary $\m D_{t-1}$, we define
\begin{equation}
\m H_{i,t}
=
\m D_{t-1}^{\transp}\bar{\m G}_i\m D_{t-1}
+\varepsilon\m I,
\qquad
\m R_{i,t}
=
\m D_{t-1}^{\transp}\bar{\m G}_i\m W_i.
\label{eq:htp_system}
\end{equation}
We warm-start HTP with the coefficients from the
previous outer iteration. Each inner iteration
comprises a gradient step, support selection,
and restricted least-squares refinement.\\
\textbf{Gradient Update:}
At outer iteration $t$, initialize the HTP coefficients
as $\m C_i^{(0)}=\m C_{i,t-1}$.
For inner iteration $h$, compute
\begin{equation}
\m C_i^{\mathrm{tmp},(h)}
=
\m C_i^{(h)}
+
\left(
\m R_{i,t}
-
\m H_{i,t}\m C_i^{(h)}
\right)
\label{eq:htp_gradient}
\end{equation}
\textbf{Support Selection:}
In each column of $\m C_i^{\mathrm{tmp},(h)}$,
select the $k$ entries with the largest absolute values.
Define the binary mask
$\m M_i^{(h+1)}\in\{0,1\}^{r\times d_i}$
to be one at the selected locations and zero elsewhere.\\

\textbf{Restricted Projection:}
With the support fixed, compute $\m C_i^{(h+1)}$
by approximately solving
\begin{align}
\min_{\m C}\quad
&
\norm{
\bar{\m L}_i(\m W_i-\m D_{t-1}\m C)
}{F}^{2}
+\varepsilon\norm{\m C}{F}^{2}
\nonumber\\
\text{s.t.}\quad
&
\m C=\m C\odot\m M_i^{(h+1)}.
\label{eq:htp_restricted}
\end{align}
The restricted normal equations are
\begin{equation}
\m M_i^{(h+1)}
\odot
(\m H_{i,t}\m C-\m R_{i,t})
=\m 0,
\end{equation}
with coefficients outside the selected support fixed
to zero. We approximately solve this system using
batched Conjugate Gradient (CG), initialized with
$\m C_i^{(h)}\odot\m M_i^{(h+1)}$.

After $T_{\mathrm{HTP}}$ inner iterations, we set
$\m C_{i,t}=\m C_i^{(T_{\mathrm{HTP}})}$
and proceed to the dictionary update.
We use two HTP iterations and at most ten CG iterations
per restricted solve.
When $k=r$, the sparsity constraint is inactive,
and we use the dense coefficient update in
Eq.~\eqref{eq:coefficient_update}.

The finite-step HTP+CG procedure constitutes an inexact
coefficient update. Convergence results based on
BSUM and Kurdyka--\L{}ojasiewicz theory require
additional conditions, including suitable decrease
and control of the inexact updates.
We do not establish these conditions for the implemented
finite-step solver. Its practical behavior is assessed
through the iteration-budget ablation.
Further discussion is provided in
Appendix~\ref{convergence_pr}.

\subsection{Cross-Layer Grouping via Graph Matching}
\label{sec:grouping}

While prior compression pipelines default to pairing adjacent layers, structural similarities in pretrained weight matrices are not strictly localized. To maximize the efficacy of cross-layer dictionary sharing, we formulate layer grouping as a global optimization problem that pairs weights with minimal structural discrepancy.

\paragraph{Whitened-Space Frobenius Submatrix Distance}
 Given two projection matrices $\m W_i \in \mathbb{R}^{d \times d_i}$ and $\m W_j \in \mathbb{R}^{d \times d_j}$ sharing the same input dimension $d$ but potentially differing in output dimension, we define a scale-invariant surrogate metric for the joint approximation error. Without loss of generality, assume $d_i < d_j$. The normalized Frobenius submatrix distance is computed in the calibration-induced whitened space as:
\begin{equation}
\delta(\m W_i , \m W_j) = \min_{0\le k\le d_j - d_i} \frac{\|\m L_j \m W_j[:,k:k+d_i] - \m L_i \m W_i \|_F}{\| \m L_i \m W_i \|_F + \epsilon}, 
\end{equation}

For the case of $d_i = d_j$, please refer to appendix \ref{impl_details}.

where $\m L_i, \m L_j$ are the layer-specific Cholesky whitening transforms derived from calibration activations, $\m W_j[:,k:k+d_i]$ extracts a contiguous column window of width $d_i$, and $\epsilon> 0$ ensures numerical stability. This metric captures the minimal alignment cost between the two weight spaces while explicitly accounting for distinct activation geometries and is computed in $\mathcal{O}\pr{d \cdot d_j}$ time using optimized 1D cross-correlation, avoiding explicit window allocations.

\paragraph{Global Maximum-Weight Matching.}
Crucially, our grouping strategy is not restricted to identical submodule types; attention and feed-forward weights can be paired whenever they share a common input dimension $d$. Let $\mathcal{V} = \{1, \dots, N\}$ index all candidate weight matrices across layers and projection types. We construct an undirected complete graph $\mathcal{G} = (\mathcal{V}, \mathcal{E})$ with edge weights defined as $w_{ij} = \mathcal{C} - \delta\pr{\m W_i, \m W_j}$, where $\mathcal{C} > \max_{i,j} \delta\pr{\m W_i, \m W_j}$ converts distance minimization into weight maximization. The pairing $\mathcal{P}^*$ is obtained by solving:
\bal
\mathcal{P}^* = \argmax_{\mathcal{M} \subseteq \mathcal{E}} \sum_{\{i,j\} \in \mathcal{M}} w_{ij} \quad \text{s.t.} \quad \mathcal{M} \text{ is a valid matching},
\eal
which simultaneously enforces maximum cardinality and minimal total structural distance. We solve this problem using the Edmonds' Blossom algorithm \cite{blossom}. The resulting disjoint pairs are subsequently passed to the calibration-aware alternating minimization of Section~\ref{sec:paired_objective}, ensuring that dictionary sharing is restricted to structurally aligned projections rather than arbitrary adjacent layers. This data-driven grouping strategy consistently yields lower activation-weighted reconstruction error and improved downstream perplexity compared to fixed adjacent pairing.

While the Blossom algorithm optimizes this tractable surrogate metric rather than the final joint reconstruction loss directly, we empirically verify that the whitened submatrix alignment cost is predictive of the optimized pairwise reconstruction error (see Appendix~\ref{app:surrogate_correlation}).

\subsection{Algorithmic Summary}
\label{sec:method_summary}
For clarity, we consolidate the complete alternating minimization procedure into its explicit initialization and iterative update rules:
\begin{itemize}
\item \textbf{Initialization ($t=0$):}
Let
$[\m W_1\;\m W_2]=\m U\bm\Sigma\m V^{\transp}$
be a singular value decomposition, with singular
values ordered nonincreasingly. Initialize
\begin{equation}
\label{eq:initialization}
\m D_0 = \m U_{:,1\!:\!r}.
\end{equation}
For the sparse variant, we initialize $\m C_{i,0}$ using
Eq.~\eqref{eq:coefficient_update} with $\m D_{t-1}=\m D_0$.
    
\item \textbf{Coefficient Update ($t\geq1$):}
Using $\m H_{i,t}$ and $\m R_{i,t}$ from
Eq.~\eqref{eq:htp_system}, compute
\begin{equation}
\m C_{i,t}
=
\begin{cases}
\operatorname{HTP}
\left(
\m H_{i,t},\m R_{i,t},\m C_{i,t-1};k
\right),
& k<r,\\[2mm]
\m H_{i,t}^{-1}\m R_{i,t},
& k=r.
\end{cases}
\label{eq:coefficient_summary}
\end{equation}
Here $\operatorname{HTP}$ denotes the finite-budget
HTP+CG procedure described in
Section~\ref{sec:htp_coefficients}, initialized with
$\m C_{i,t-1}$.
    
\item \textbf{Dictionary Update ($t\geq1$):}
\bal
\m D_t
&= \argmin_{\m D}
\sum_{i=1}^{2}
\left[
\norm{\bar{\m L}_i(\m W_i-\m D\m C_{i,t})}{F}^{2}
+\epsilon\norm{\bar{\m L}_i\m D}{F}^{2}
\right]
\nonumber\\
&= \m P
\left(
\frac{\m P^\transp\m K_t\m Q_t}
{\mathbf{1}_d\mathbf{1}_r^\transp
+\bm\lambda\bm\sigma_t^\transp}
\right)
\m Q_t^\transp.
\eal
    
\end{itemize}
We run at most 50 outer iterations and stop after
three consecutive iterations that fail to improve
the best recorded reconstruction loss by more than
$10^{-5}$. We return the factors attaining the lowest
observed reconstruction loss.

For practical storage details and throughput analysis, please refer to the appendix \ref{app:macko_serving}.
\section{Experiments}
This section systematically evaluates our approach, hereafter referred to as \ours, across design components and assess performance across  different settings using 7 well established benchmarks (we refer to Appendix \ref{impl_details} for details). We begin with a component-wise ablation on two representative language models, comparing each configuration against the Basis Sharing baseline to isolate the impact of our proposed modules. Following this analysis, we benchmark our method against recent dictionary learning approaches to establish its effectiveness within this paradigm. We then evaluate the framework against a broad set of pruning and compression techniques, demonstrating that our training-free pipeline achieves competitive accuracy without the post-compression fine-tuning typically required by existing methods. To further assess scalability and architectural robustness, we extend the comparison against Basis Sharing across varying compression ratios and diverse model families. Finally, we apply the framework to a recent video generation model, providing qualitative evidence that high-fidelity generation is preserved without any post-compression adaptation or recovery steps.

\paragraph{Pairwise Weight Optimization and Coefficient Sparsification}
Table~\ref{tab:compression_results} evaluates the
Sylvester update, grouping strategy, and sparse
coefficient representation at $\mathrm{CR}=0.2$.
On Llama-3.2 1B, replacing global whitening with
the Sylvester update increases average accuracy
from 33.11\% to 41.26\%, and incorporating the
proposed grouping increases it further to 43.03\%.
The corresponding results on Llama-3 8B are
53.47\%, 56.40\%, and 57.46\%.

Moving from dense factorization to sparse dictionary
factorization produces the largest accuracy increase.
This comparison changes the representation family;
it does not isolate the benefit of HTP itself.
Within the sparse representation, the proposed
grouping improves average accuracy from 53.06\%
to 54.09\% on Llama-3.2 1B and from 67.31\%
to 68.17\% on Llama-3 8B.
Thus, the geometry-preserving dictionary update
and grouping provide improvements within the
evaluated representation families.
Additional comparisons with Basis Sharing and
CoSpaDi are provided in Appendix~\ref{comp_bs_cos}.

\begin{table}[h]
\centering
\caption{Component ablation at $\mathrm{CR}=0.2$.
GW denotes global whitening; Sylv denotes the
geometry-preserving Sylvester update; OG denotes
our graph-matching strategy; and SP denotes sparse
coefficients obtained using HTP.
Dense and sparse configurations use our Sylvester
update with either adjacent or matched pairing.}
\label{tab:compression_results}

\resizebox{\textwidth}{!}{%
\begin{tabular}{lcccccccccccccccc}
\toprule
\multirow{2}{*}{Configuration}
& \multicolumn{4}{c}{Components}
& \multirow{2}{*}{CR}
& \multicolumn{8}{c}{Benchmarks (Accuracy)}
& \multicolumn{2}{c}{Perplexity}
& \multirow{2}{*}{\makecell{Avg\\Acc}} \\
\cmidrule(lr){2-5}
\cmidrule(lr){7-14}
\cmidrule(lr){15-16}
& GW & Sylv & OG & SP &
& PIQA & HellaSwag & Lambada\_OA & ARC-e & ARC-c
& SciQ & Race & MMLU & Wiki & Lambada & \\
\midrule

Llama3.2 1B (baseline)
& \xmark & \xmark & \xmark & \xmark & --
& 74.53 & 63.66 & 62.95 & 60.47 & 36.20 & 88.30
& 37.79 & 37.00 & 11.60 & 5.73 & 57.61 \\

Basis Sharing
& \cmark & \xmark & \xmark & \xmark & 0.2
& 56.75 & 31.69 & 15.08 & 32.49 & 21.59 & 58.40
& 25.93 & 22.95 & 928.07 & 239.30 & 33.11 \\

Dense Coeff. -- adjacent
& \xmark & \cmark & \xmark & \xmark & 0.2
& 63.76 & 40.21 & 32.91 & 41.33 & 24.74 & 73.80
& 29.86 & 23.44 & 109.59 & 57.95 & 41.26 \\

Dense Coeff. -- matched
& \xmark & \cmark & \cmark & \xmark & 0.2
& 64.91 & 42.64 & 37.42 & 44.02 & 25.94 & 76.30
& 29.95 & 23.05 & 56.60 & 32.49 & 43.03 \\

Sparse Coeff. -- adjacent
& \xmark & \cmark & \xmark & \cmark & 0.2
& 71.44 & 57.35 & 56.55 & 54.46 & 32.51 & 87.50
& 34.93 & 29.70 & 16.77 & 7.88 & 53.06 \\

Sparse Coeff. -- matched
& \xmark & \cmark & \cmark & \cmark & 0.2
& 73.50 & 57.67 & 58.88 & 57.74 & 31.83 & 88.90
& 34.26 & 29.96
& \textbf{15.92} & \textbf{6.74} & \textbf{54.09} \\

\midrule

Llama3 8B (baseline)
& \xmark & \xmark & \xmark & \xmark & --
& 80.69 & 79.13 & 75.57 & 77.69 & 53.5 & 93.9
& 40.29 & 62.15 & 7.26 & 3.09 & 70.36 \\

Basis Sharing
& \cmark & \xmark & \xmark & \xmark & 0.2
& 72.52 & 58.71 & 50.2 & 57.2 & 34.04 & 84.9
& 37.13 & 33.04 & 41.26 & 14.84 & 53.47 \\

Dense Coeff. -- adjacent
& \xmark & \cmark & \xmark & \xmark & 0.2
& 74.21 & 61.14 & 57.4 & 59.76 & 36.18 & 86.9
& 37.13 & 38.47 & 26.82 & 9.69 & 56.4 \\

Dense Coeff. -- matched
& \xmark & \cmark & \cmark & \xmark & 0.2
& 73.5 & 59.8 & 60.7 & 65.11 & 37.29 & 90.8
& 37.42 & 35.09 & 24.56 & 6.82 & 57.46 \\

Sparse Coeff. -- adjacent
& \xmark & \cmark & \xmark & \cmark & 0.2
& 78.35 & 76.18 & 70.62 & 73.4 & 49.91 & 93.2
& 39.43 & 57.37 & 9.75 & 4.15 & 67.31 \\

Sparse Coeff. -- matched
& \xmark & \cmark & \cmark & \cmark & 0.2
& 78.73 & 75.38 & 74.25 & 76.14 & 49.74 & 93.9
& 40.77 & 56.45
& \textbf{9.43} & \textbf{3.25} & \textbf{68.17} \\

\bottomrule
\end{tabular}%
}
\end{table}

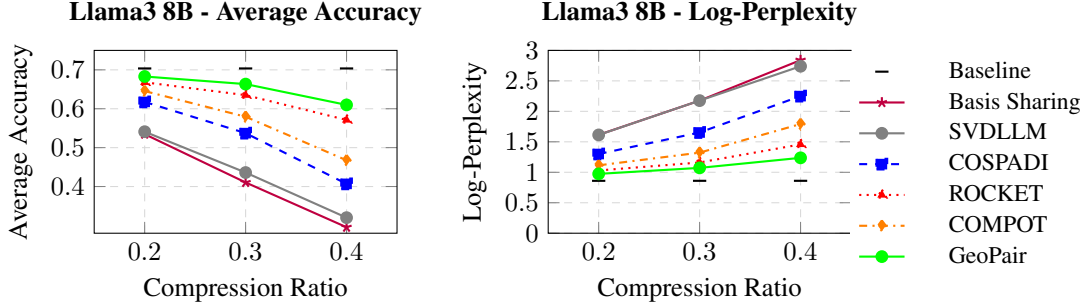
\begin{figure}[ht]
    \begin{tikzpicture}
    \begin{groupplot}[
        group style={
            group size=2 by 1,
            horizontal sep=2cm,
            xlabels at=edge bottom,
            ylabels at=edge left
        },
        width=0.4\textwidth,
        height=4cm,
        grid=major,
        grid style={dashed,gray!30},
        legend style={
            font=\small,
            cells={anchor=west},
            legend columns=1,
            column sep=0.5cm,
            at={(1.4,1.0)},
            anchor=north,
            draw=none,
        },
        title style={font=\bfseries, align=center},
        every axis plot/.append style={thick},
        xmin=0.15, xmax=0.45,
        xtick={0.2, 0.3, 0.4},
    ]

    \nextgroupplot[
        title={Llama3 8B - Average Accuracy},
        ylabel={Average Accuracy},
        xlabel={Compression Ratio},
        ymin=0.28, ymax=0.75,
        ytick={0.4, 0.5, 0.6, 0.7},
    ]
    \addplot[black, mark=-, only marks, mark size=2.5pt] coordinates {(0,0.7036)(0.2,0.7036)(0.3,0.7036)(0.4,0.7036)};
    \addplot[purple, thick, mark=star] coordinates {(0.2,0.5347)(0.3,0.4105)(0.4,0.2954)};
    \addplot[gray, thick, mark=otimes*] coordinates {(0.2,0.541)(0.3,0.436)(0.4,0.32)};
    \addplot[blue, dashed, mark=square*] coordinates {(0.2,0.6176)(0.3,0.5373)(0.4,0.4069)};
    \addplot[red, dotted, mark=triangle*] coordinates {(0.2,0.6680)(0.3,0.6350)(0.4,0.5712)};
    \addplot[orange, dashdotted, mark=diamond*] coordinates {(0.2,0.6454)(0.3,0.5794)(0.4,0.4675)};
    \addplot[green, thick, mark=otimes*] coordinates {(0.2,0.6829)(0.3,0.6634)(0.4,0.6099)};


    \nextgroupplot[
        title={Llama3 8B - Log-Perplexity},
        ylabel={Log-Perplexity},
        xlabel={Compression Ratio},
        ymin=0, ymax=3,
        ytick={0, 0.5, 1.0, 1.5, 2.0, 2.5, 3.0},
    ]
    \addplot[black, mark=-, only marks, mark size=2.5pt] coordinates {(0.2,0.86)(0.3,0.86)(0.4,0.86)};
    \addplot[purple, thick, mark=star] coordinates {(0.2,1.615)(0.3,2.176)(0.4,2.84)};
    \addplot[gray, thick, mark=otimes*] coordinates {(0.2,1.613)(0.3,2.176)(0.4,2.74)};
    \addplot[blue, dashed, mark=square*] coordinates {(0.2,1.3)(0.3,1.653)(0.4,2.25)};
    \addplot[red, dotted, mark=triangle*] coordinates {(0.2,1.03)(0.3,1.161)(0.4,1.4577)};
    \addplot[orange, dashdotted, mark=diamond*] coordinates {(0.2,1.1139)(0.3,1.3222)(0.4,1.792)};
    \addplot[green, thick, mark=otimes*] coordinates {(0.2,0.974)(0.3,1.0722)(0.4,1.237)};



    \legend{Baseline,  Basis Sharing, SVDLLM, COSPADI, ROCKET, COMPOT, \ours{}}

    \end{groupplot}
    \end{tikzpicture}\vspace{-0.2cm}
    \caption{Comparison of different compression methods on Llama3 8B across varying compression ratios. Left: Average accuracy across benchmarks. Right: Log-Perplexity on WikiText.} 
    \label{fig:compression_comparison}
    \vspace{-0.5cm}
\end{figure}

\paragraph{Shared Dictionary Learns Better}
Figure~\ref{fig:compression_comparison} evaluates \ours{} against a broad set of structured weight factorization methods, encompassing both dense low-rank projections and sparse dictionary learning approaches. \ours{} consistently achieves the highest accuracy, outperforming established baselines in both categories. This advantage stems from our shared dictionary formulation and grouping strategy, which more effectively preserves weight structure than conventional low-rank approximations or other dictionary learning strategies. These results establish \ours{} as the leading training-free weight factorization method for the evaluated compression regime.

\paragraph{Comparison with other compression methods}
Table~\ref{tab:pruning_comparison} evaluates \ours{} against a broad range of compression and pruning strategies, including methods that operate outside the matrix factorization paradigm. Under a consistent evaluation protocol, \ours{} achieves the highest average accuracy across all benchmarks while operating entirely without post-compression fine-tuning. In contrast, competing approaches rely on extensive healing or recovery training, which introduces significant computational overhead and requires large datasets. These results establish \ours{} as the state-of-the-art training-free compression method.
\begin{table*}[h]
\centering
\small
\caption{Comparison against pruning  methods on Llama2 7B across standard zero-shot benchmarks at 20\% compression. Training-free indicates whether a finetuning is done after compression.}
\resizebox{0.8\textwidth}{!}{
\begin{tabular}{lcccccccccc}
\toprule
\multirow{2}{*}{Method} & \multirow{2}{*}{Training-free} & \multicolumn{7}{c}{Accuracy$\uparrow$} & \multirow{2}{*}{Avg.} \\
\cmidrule(lr){3-9}
& & BoolQ & PIQA & HellaSwag & WinoGrande & ARC-e & ARC-c & OBQA & \\
\midrule
Baseline & -- & 76.50 & 79.80 & 76.10 & 70.10 & 72.80 & 47.60 & 57.20 & 68.59 \\
\midrule
LLM-Pruner & \ding{55} & 66.79 & 77.58 & 68.48 & 64.96 & 64.06 & 37.88 & 39.00 & 59.82 \\
LoRAPrune & \ding{55} & 65.82 & \textbf{79.31} & 70.00 & 62.76 & 65.87 & 37.69 & 39.14 & 60.05 \\
WANDA & \ding{51} & 65.75 & 74.70 & 64.52 & 59.35 & 60.65 & 36.26 & 39.40 & 57.23 \\
ShortGPT & \ding{55} & 68.26 & 72.28 & 61.70 & 63.77 & 60.22 & 39.00 & 41.60 & 58.12 \\
LoRAShear & \ding{55} & 72.78 & 76.36 & 69.49 & \textbf{67.63} & 69.02 & 39.47 & 40.78 & 62.22 \\
\ours & \ding{51} & \textbf{74.06} & 77.2 & \textbf{71.8} & 67.24 & \textbf{72.1} & \textbf{41.3} & \textbf{41.4} & \textbf{63.58} \\
\bottomrule
\end{tabular}
}

\label{tab:pruning_comparison}
\vspace{-0.3 cm}
\end{table*}

\paragraph{Generalization Across Model Architectures}
To assess cross-architecture robustness, we evaluate our compression pipeline across diverse model families and scales, ranging from 1B to 32B parameters. This evaluation verifies that our framework maintains effectiveness irrespective of model capacity, architectural design, or training paradigm. Table~\ref{tab:compression_results_placeholder} reports average zero-shot accuracy and Lambada OpenAI perplexity under increasing compression ratios. For direct comparison, we include Basis Sharing results at compression ratios 0.2, 0.3, and 0.4, enabling a consistent assessment of performance trends across compression intensities. Additional results for Qwen3 on an updated benchmark suite are provided in Appendix~\ref{new_lb} to confirm consistency under alternative evaluation protocols.
We observe minor improvements on small compression ratios, this aligns with recent findings that small rank truncation acts as denoising, preserving salient features~\cite{laser}.

\begin{table*}[h]
\centering
\small
\caption{Generalization across model families and parameter scales. For each architecture, we report average zero-shot accuracy and Lambada OpenAI perplexity under increasing compression ratios. At CR=0.2, 0.3, and 0.4, we include direct comparisons against Basis Sharing.}
\resizebox{0.8\textwidth}{!}{%
\begin{tabular}{lc|c|cc|cc|cc|cc}
\toprule
\multirow{2}{*}{Model} & \multirow{2}{*}{Metric} & \multirow{2}{*}{CR=0} & \multicolumn{2}{c}{CR=0.2} & \multicolumn{2}{c}{CR=0.3} & \multicolumn{2}{c}{CR=0.4} \\
\cmidrule(lr){4-5} \cmidrule(lr){6-7} \cmidrule(lr){8-9}
& & & Ours & Basis Sharing & Ours & Basis Sharing & Ours & Basis Sharing \\
\midrule
\multirow{2}{*}{Qwen 3 8B} 
& Avg. Accuracy & 70.46 & \textbf{67.7} & 61.25 & \textbf{64.3} & 55.17 & \textbf{58.88} & 46.41 \\
& Perplexity & 4.60 & \textbf{4.87} & 7.31 & \textbf{6.56} & 12.94 & \textbf{13.40} & 45.21 \\
\midrule
\multirow{2}{*}{Gemma 3 12B} 
& Avg. Accuracy & 72.28 & \textbf{71.64} & 58.96 & \textbf{67.72} & 50.2 & \textbf{60.49} & 40.22 \\
& Perplexity & 4.16 & \textbf{3.62} & 35.65 & \textbf{4.39} & 164.3 & \textbf{10.98} & 765.7 \\
\midrule
\multirow{2}{*}{Phi-4 14B} 
& Avg. Accuracy & 72.09 & \textbf{74.07} & 70.33 & \textbf{70.25} & 65.22 & \textbf{68.59} & 57.52 \\
& Perplexity & 3.49 & \textbf{3.37} & 3.44 & \textbf{3.80}& 4.33 & \textbf{4.35} & 8.3 \\
\midrule
\multirow{2}{*}{Qwen 3 32B} 
& Avg. Accuracy & 74.55 & \textbf{73.27} & 69.64& \textbf{72.24} & 66.04 & \textbf{70.31} & 59.5 \\
& Perplexity & 3.74 & \textbf{3.26} & 3.51 & \textbf{3.22} & 4.09 & \textbf{3.38} & 6.6\\
\bottomrule
\end{tabular}%
}

\label{tab:compression_results_placeholder}
\vspace{-0.2 cm}
\end{table*}

\subsection{Generalization across tasks}
To evaluate cross-task generalization, we apply \ours{} to a video generation model. Specifically, we compress Wan2.2 5B \cite{wan2025wan0} at 20\% and 40\% compression ratios and demonstrate that the compressed models retain high-fidelity video generation capabilities without any post-compression fine-tuning. To quantitatively assess the preservation of semantic alignment under compression, we evaluate generated videos using X-CLIP~\cite{xclip} with 16-frame sampling over 50 prompts drawn from the \texttt{Rapidata/awesome-text2video-prompts} dataset. The compressed models exhibit near-baseline performance: at 20\% compression, the average CLIP score decreases by only $1 \times 10^{-4}$ (0.2164 vs.~0.2165 baseline), while even at 40\% compression the relative degradation remains marginal (0.2111, a 2.5\% drop). These results confirm that our training-free compression framework preserves cross-modal alignment and generalization capacity without task-specific adaptation or recovery procedures. Compression results on audio generation models are provided in Appendix~\ref{audio}.

\begin{figure}[h]
    \centering
    \includegraphics[width=0.9\linewidth]{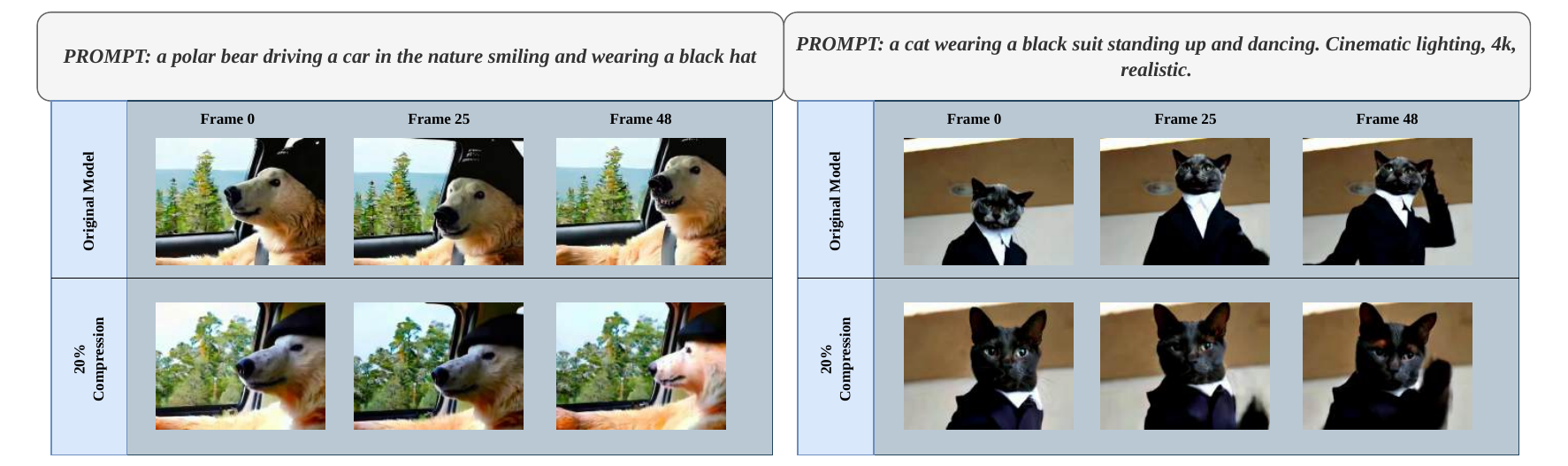}
    \caption{Video frames generated using Wan2.2 5B model and a compressed version using \ours{}.}
    \label{fig:share-flower-qr}
    \vspace{-0.3 cm}
\end{figure}

\section{Ablation Studies}
This section evaluates the core components of the proposed compression framework to validate key design choices. We begin by analyzing the layer grouping strategy, comparing alternative distance metrics to determine the most effective criterion for pairing weight matrices. Next, we examine the framework's sensitivity to the KS ratio (defined as the number of atoms in the dictionary over the non-zero elements in each column of the coefficient matrices $\m C_i$). In Appendix \ref{extra_ablation} we benchmark alternative sparse algorithms for enforcing target sparsity levels and perform  ablations related to \ours's convergence and sensitivity on calibration data.
\paragraph{Grouping algorithm}
The proposed grouping strategy is optimal with respect to a chosen distance metric for each pair of weights, 
In Table \ref{tab:grouping_ablation}, we ablate two different metrics as well as the greedy approach used in Basis Sharing~\cite{basis_sharing}. 
\begin{table}[h]

\centering
\small
\caption{Ablation study of layer grouping strategies at CR=0.4 and KS=2.5 on Llama3.2 1B. Frobenius norm-based grouping achieves the best trade-off.}
\begin{tabular}{lccc}
\toprule
\multirow{2}{*}{Grouping Strategy} & \multirow{2}{*}{CR} & \multicolumn{1}{c}{WikiText-2} & \multirow{2}{*}{Avg. Accuracy} \\
& & \multicolumn{1}{c}{(Perplexity$\downarrow$)} & \\
\midrule
Baseline (Llama3.2 1B) & -- & 11.60 & 57.6 \\
\midrule
General Frobenius Norm & 0.4 & \textbf{34.26} & \textbf{45.61} \\
Cosine Similarity & 0.4 & 43.75 & 45.0 \\
Consecutive Layers (Greedy) & 0.4 & 37.97 & 43.35 \\
\bottomrule
\end{tabular}

\label{tab:grouping_ablation}
\vspace{-0.2 cm}
\end{table}

\paragraph{Ablation on the KS ratio}
From Table \ref{tab:ks_ratio_ablation} we observe that KS equal to 2.5 leads to the best results. Based on this observation, we set ${k/s} =2.5$ for all the reported experiments that include sparsification of matrix coefficients. 

\begin{table}[h]

\centering
\small
\caption{Ablation study of the KS ratio at CR=0.4 on Llama3.2 1B. The KS ratio balances the sparsity level of the coefficients and the rank of the dictionary. Lower perplexity and higher accuracy indicate better preservation of model capability. Intermediate KS values (2.5--3.5) yield the most favorable trade-offs, with KS=2.5 selected as the default configuration.}
\begin{tabular}{lcccccc}
\toprule
\multirow{2}{*}{Metric} & \multirow{2}{*}{Baseline} & \multicolumn{4}{c}{KS Ratio (CR=0.4)} \\
\cmidrule(lr){3-6}
& & 2.0 & 2.5 & 3.0 & 3.5 & 4.0 \\
\midrule
WikiText-2 (Perplexity$\downarrow$) & 11.60 & 48.51 & 34.26 & 33.96 & \textbf{33.53} & 35.43 \\
Lambada (Perplexity$\downarrow$)    & 5.73  & 27.60 & \textbf{21.69} & 21.80 & 22.71 & 24.55 \\
Avg. Accuracy ($\uparrow$)          & 57.6  & 43.5  & \textbf{45.6}  & 45.1  & 45.0  & 44.3  \\
\bottomrule
\end{tabular}

\label{tab:ks_ratio_ablation}
\end{table}
\vspace{-0.3cm}
\section{Conclusion \& Limitations}
\ours{} introduces a training-free framework for
pairwise dictionary sharing under layer-specific
calibration geometries.
The framework combines closed-form regularized
Sylvester dictionary updates, maximum-weight matching
under the proposed surrogate cost, and optional
HTP-based sparse coefficient optimization.
Experiments demonstrate favorable accuracy/compression
trade-offs across the evaluated language models and
show applicability to video and audio generation,
without post-compression fine-tuning.

A primary limitation of the current framework is its restriction to pairwise layer grouping. Extending the optimization to simultaneously share dictionaries across larger groups of layers ($m > 2$) is non-trivial, as the Sylvester solver relies on the simultaneous diagonalization of two matrix pencils, a property that does not analytically generalize to $m$-term systems. Multi-group extensions would necessitate iterative numerical approximations or higher-order tensor factorizations, introducing additional computational overhead and potential numerical instability. Developing scalable, theoretically grounded strategies for $m$-way layer sharing remains a key challenge and constitutes a primary direction for future work.

\paragraph{Acknowledgments}
This research was supported as part of the ITMO University Research Projects in AI Initiative.

\newpage
\bibliography{neurips_2026_1}
\bibliographystyle{acm}

\appendix

\newpage
\section{Theoretical background}
\label{full_theory}
\subsection{Background and recap}
\label{sec:background}
\paragraph{Sylvester Equation.}
The classical Sylvester equation is a linear matrix equation of the form
\bal
\m A \m X + \m X \m B = \m C,
\eal
where $\m A \in \R^{m \times m}$, $\m B \in \R^{n \times n}$, and $\m C \in \R^{m \times n}$ are given matrices, and $\m X \in \R^{m \times n}$ is the unknown. This equation arises frequently in control theory, model order reduction, and structured matrix factorization. A unique solution exists if and only if the spectra of $\m A$ and $-\m B$ are disjoint, i.e., $\lambda_i(\m A) + \lambda_j(\m B) \neq 0$ for all eigenvalue pairs $(\lambda_i(\m A), \lambda_j(\m B))$ \cite{Sylvester1884}. While the system can be vectorized using Kronecker products as $(\m I_n \otimes \m A + \m B^\transp \otimes \m I_m)\mathrm{vec}(\m X) = \mathrm{vec}(\m C)$, direct inversion scales poorly ($\mathcal{O}(m^3 n^3)$) and is numerically unstable for large dimensions. Instead, the Bartels-Stewart algorithm \cite{Bartels1972} exploits Schur decompositions to solve the system efficiently in $\mathcal{O}(m^3 + n^3)$ time with guaranteed numerical stability.

\paragraph{Generalized Sylvester Equation.}
In our dictionary update step, we encounter a \emph{two-term generalized Sylvester equation} of the form
\bal
\m A_1 \m X \m B_1 + \m A_2 \m X \m B_2 = \m C.
\label{eq:gen_sylvester}
\eal
More generally, an $m$-term generalized Sylvester equation takes the form $\sum_{k=1}^m \m A_k \m X \m B_k = \m C$. Unlike the classical case, no universal closed-form solution exists, and the feasibility of the solution depends on the spectral properties of the matrix pencils $(\m A_1, \m A_2)$ and $(\m B_1, \m B_2)$. However, when $\mathbf{A}_1$ and $\mathbf{B}_1$ are symmetric positive definite and $\mathbf{A}_2, \mathbf{B}_2$ are symmetric positive semi-definite, the system can be decoupled via simultaneous diagonalization. In our case, strict positive definiteness of $\mathbf{B}_1$ is enforced via Tikhonov regularization ($\tilde{\mathbf{B}}_1 = \mathbf{B}_1 + \epsilon \mathbf{I}$) to ensure the generalized eigenvalue decomposition is well-posed. Specifically, we compute the generalized eigenvalue decompositions (GEVD) $(\m P, \bm\Lambda) = \Phi(\m A_2, \m A_1)$ and $(\m Q, \bm\Sigma) = \Phi(\m B_2, \m B_1)$, which satisfy
\bal
\m P^\transp \m A_1 \m P = \m I, \quad \m P^\transp \m A_2 \m P = \bm\Lambda, \qquad
\m Q^\transp \m B_1 \m Q = \m I, \quad \m Q^\transp \m B_2 \m Q = \bm\Sigma.
\eal
Substituting $\m X = \m P \m Y \m Q^T$ into Eq.(13) transforms the coupled matrix system into a set of independent scalar equations:
\begin{equation}
(1 + \lambda_i \sigma_j) y_{ij} = (\m P^T \m C \m Q)_{ij}, \quad \forall i,j, 
\end{equation}
where $\lambda_i$ and $\sigma_j$ denote the diagonal entries of $\bm \Lambda$ and $\bm\Sigma$, respectively. Provided that $1 + \lambda_i \sigma_j \neq 0$ for all $i,j$  a condition naturally satisfied in our setting since all eigenvalues are non-negative the solution admits a closed-form expression via element-wise division:
\begin{equation}
\m Y = \frac{\m P^T \m C \m Q}{\mathbf{1}\mathbf{1}^T + \bm\lambda \bm\sigma^T}, 
\end{equation}
where $\bm \lambda = \text{diag}(\bm \Lambda)$ and $\bm \sigma = \text{diag}(\bm \Sigma)$ are eigenvalue vectors, and the denominator represents the outer product structure for proper broadcasting.
The final solution is recovered as $\m X = \m P \m Y \m Q^\transp$. This simultaneous diagonalization strategy avoids iterative solvers, eliminates step-size tuning, and guarantees deterministic convergence with $\mathcal{O}(d^3)$ complexity dominated by the initial GEVD computation.

For the dictionary update in
Section~\ref{sec:paired_objective}, identify
\[
\m A_i=\bar{\m G}_i,
\qquad
\m B_i=\tilde{\m B}_{i,t}
=\m C_{i,t}\m C_{i,t}^{\transp}+\epsilon\m I.
\]
The normalized activation Gram matrices
$\bar{\m G}_i$ are positive definite for
$\eta>0$ and $s_i>0$, while
$\tilde{\m B}_{i,t}$ are positive definite for
$\epsilon>0$.
The generalized eigenvalues are therefore positive,
and $1+\lambda_a\sigma_b>0$ for all $a,b$.
The resulting solution is the unique minimizer
of the regularized dictionary subproblem.
\paragraph{Hard Thresholding Pursuit.}
Hard Thresholding Pursuit (HTP)~\cite{htp} is a greedy iterative algorithm designed to solve $\ell_0$-constrained least squares problems, widely adopted in compressive sensing and sparse dictionary learning. Unlike basic thresholding schemes that merely zero out non-dominant coefficients, HTP enhances reconstruction accuracy by coupling a gradient-based support identification step with a least-squares projection onto the selected subspace. Given an overcomplete system $\m y = \bm \Phi \m c + \bm \epsilon$, HTP seeks a $k$-sparse solution through the following canonical iteration at step $j$:
\bal
\m c^{\text{tmp}} &= \m c^{(j)} + \mu \bm \Phi^\transp \pr{\m y - \bm \Phi \m c^{(j)}}, \quad \m S^{(j+1)} = \operatorname{supp}\!\big(\mathcal{H}_k(\m c^{\text{tmp}})\big), \\
\m c^{(j+1)} &= \argmin_{\m c \, : \, \operatorname{supp}(\m c) \subseteq \m S^{(j+1)}} \norm{\m y - \bm \Phi \m c}{2}^2,
\eal
where $\mathcal{H}_k(\cdot)$ retains the $k$ entries of largest magnitude, $\operatorname{supp}(\cdot)$ extracts the corresponding index set, and $\mu > 0$ is a step size. This restricted projection ensures that the active coefficients are optimally fitted to the measurements, yielding superior convergence and fidelity over pure iterative thresholding.

For a fixed dictionary $\m D_{t-1}$, the coefficient
subproblem is
\begin{align}
\min_{\m C}\quad
&
\norm{
\bar{\m L}_i(\m W_i-\m D_{t-1}\m C)
}{F}^{2}
+\varepsilon\norm{\m C}{F}^{2}
\nonumber\\
\text{s.t.}\quad
&
\norm{\m C[:,j]}{0}\leq k,
\qquad j=1,\ldots,d_i.
\label{eq:htp_constraint}
\end{align}
Define
\[
\m H_{i,t}
=\m D_{t-1}^{\transp}\bar{\m G}_i\m D_{t-1}
+\varepsilon\m I,
\qquad
\m R_{i,t}
=\m D_{t-1}^{\transp}\bar{\m G}_i\m W_i.
\]
Initialize $\m C_i^{(0)}=\m C_{i,t-1}$.
Each HTP inner iteration consists of:
\begin{enumerate}
\item \textbf{Gradient Update:}
\[
\m C_i^{\mathrm{tmp},(h)}
=
\m C_i^{(h)}
+\mu(\m R_{i,t}-\m H_{i,t}\m C_i^{(h)}),
\qquad \mu=1.
\]

\item \textbf{Support Selection:}
Select the $k$ largest-magnitude entries in each
column of $\m C_i^{\mathrm{tmp},(h)}$ to form
$\m M_i^{(h+1)}\in\{0,1\}^{r\times d_i}$.

\item \textbf{Restricted Projection:}
Approximately solve
\[
\m M_i^{(h+1)}
\odot(\m H_{i,t}\m C-\m R_{i,t})=\m 0,
\qquad
\m C=\m C\odot\m M_i^{(h+1)},
\]
using batched CG initialized with
$\m C_i^{(h)}\odot\m M_i^{(h+1)}$.
The resulting coefficients define $\m C_i^{(h+1)}$.
\end{enumerate}
After $T_{\mathrm{HTP}}$ iterations, set
$\m C_{i,t}=\m C_i^{(T_{\mathrm{HTP}})}$.
We use two HTP iterations and at most ten CG
iterations per restricted solve.
The scope of the convergence claims is discussed
in Appendix~\ref{convergence_pr}.

\subsection{Shared Dictionary Optimization}
\label{full_sd_opt}
\paragraph{Optimization Objective.}
With the coefficients fixed, consider
\begin{equation}
J(\m D)
=
\sum_{i=1}^{2}
\left[
\norm{
\bar{\m L}_i(\m W_i-\m D\m C_{i,t})
}{F}^{2}
+
\epsilon\norm{\bar{\m L}_i\m D}{F}^{2}
\right].
\end{equation}
Define
\[
\bar{\m G}_i
=\bar{\m L}_i^\transp\bar{\m L}_i,
\qquad
\tilde{\m B}_{i,t}
=\m C_{i,t}\m C_{i,t}^\transp+\epsilon\m I.
\]
Differentiating gives
\begin{equation}
\nabla_{\m D}J
=
2\sum_{i=1}^{2}
\bar{\m G}_i
\left(
\m D\tilde{\m B}_{i,t}
-\m W_i\m C_{i,t}^\transp
\right).
\end{equation}
Setting the gradient to zero yields
\begin{equation}
\bar{\m G}_1\m D_t\tilde{\m B}_{1,t}
+
\bar{\m G}_2\m D_t\tilde{\m B}_{2,t}
=
\m K_t,
\end{equation}
where
\[
\m K_t
=
\sum_{i=1}^{2}
\bar{\m G}_i\m W_i\m C_{i,t}^\transp.
\]
This is Eq.~\eqref{eq:dictionary_normal}
of the main text.

\paragraph{Generalized Sylvester Equation.}
Compute
\[
(\m P,\bm\Lambda)
=\Phi(\bar{\m G}_2,\bar{\m G}_1),
\qquad
(\m Q_t,\bm\Sigma_t)
=\Phi(\tilde{\m B}_{2,t},\tilde{\m B}_{1,t}).
\]
Writing $\m D_t=\m P\m Y_t\m Q_t^\transp$
decouples the system into
\begin{equation}
(1+\lambda_a\sigma_{t,b})(Y_t)_{ab}
=
(\m P^\transp\m K_t\m Q_t)_{ab},
\end{equation}
where
$\bm\lambda=\diag{\bm\Lambda}$ and
$\bm\sigma_t=\diag{\bm\Sigma_t}$.
Hence
\begin{equation}
\m D_t
=
\m P
\left(
\frac{\m P^\transp\m K_t\m Q_t}
{\mathbf{1}_d\mathbf{1}_r^\transp
+\bm\lambda\bm\sigma_t^\transp}
\right)
\m Q_t^\transp.
\end{equation}
The normalized activation Gram matrices are positive
definite for $\eta>0$ and $s_i>0$.
The coefficient Gram matrices
$\tilde{\m B}_{i,t}$ are positive definite for
$\epsilon>0$.
Their generalized eigenvalues are therefore positive,
so every denominator is nonzero.
The resulting update is the unique minimizer
of the regularized dictionary subproblem.
\paragraph{Convergence Scope and Practical Behavior.}
\label{convergence_pr}

The alternating procedure combines a dictionary update
with dense or sparse coefficient updates.
The dense coefficient update exactly solves its
ridge-regularized quadratic subproblem.
The sparse coefficient update uses a finite number
of HTP iterations with approximate restricted solves
via CG. Exact solutions of individual quadratic
subproblems do not imply global optimality of the
joint factorization.

\paragraph{Conditions for Convergence.}
BSUM and Kurdyka--\L{}ojasiewicz analyses provide
convergence results under suitable assumptions on
the objective and update sequence
\cite{razaviyayn2012unified,bolte2018first}.
Applying such results requires verifying conditions
including bounded iterates, sufficient decrease,
and appropriate control of inexact updates.

For fixed calibration statistics, the reconstruction
term and coefficient penalties are polynomial
functions of the factors, and the column-wise
sparsity constraints define a semi-algebraic
feasible set. This structure alone does not establish
convergence of the implemented algorithm.
In particular, an $\ell_0$ constraint does not bound
coefficient magnitudes, and normalization of the
calibration operators does not establish boundedness
of the factor iterates.

\paragraph{Finite-Step Sparse Updates.}
A fixed HTP/CG iteration budget does not by itself
guarantee sufficient decrease or the inexactness
conditions required by convergence analyses.
We do not establish these conditions for the
implemented finite-step solver.
We also do not verify RIP-type recovery assumptions
for the evolving operators
$\bar{\m L}_i\m D_{t-1}$.
Accordingly, standalone HTP recovery guarantees
\cite{htp} do not establish convergence of the
complete alternating procedure.

\paragraph{Empirical Evaluation.}
Table~\ref{tab:convergence_ablation} evaluates the
runtime and downstream performance obtained with
different HTP and CG iteration budgets.
We select two HTP iterations and at most ten CG
iterations per restricted solve as a practical
runtime--accuracy trade-off.
These measurements support the selected computational
budget; they do not constitute a proof of convergence
or optimality.

\section{Implementation Details}
\label{impl_details}

\paragraph{Budget Allocation}
\label{app:budget}

Consider a pair of weight matrices
$W_1 \in \mathbb{R}^{d \times d_1}$ and
$W_2 \in \mathbb{R}^{d \times d_2}$
sharing the input dimension $d$.
Let $D \in \mathbb{R}^{d \times r}$ be the shared dictionary and
$C_i \in \mathbb{R}^{r \times d_i}$ the coefficient matrix of $W_i$,
with $k$ non-zero entries per coefficient column. We define
\begin{equation}
\mathrm{CR}=1-\frac{S_{\mathrm{comp}}}{S_{\mathrm{raw}}},
\qquad
\mathrm{KS}=\frac{r}{k}.
\end{equation}
The raw and compressed parameter counts are
\begin{equation}
S_{\mathrm{raw}} = d(d_1+d_2),
\qquad
S_{\mathrm{comp}} = dr + k(d_1+d_2).
\end{equation}
Using $k = \lfloor r/\mathrm{KS} \rfloor$, the rank $r$ is chosen to satisfy the storage budget
\begin{equation}
dr + \left\lfloor \frac{r}{\mathrm{KS}} \right\rfloor (d_1+d_2)
\le
(1-\mathrm{CR})\,d(d_1+d_2).
\end{equation}
Equivalently, using the continuous relaxation $k \approx r/\mathrm{KS}$, we set
\begin{equation}
r =
\left\lfloor
\frac{(1-\mathrm{CR})\,d(d_1+d_2)}
{d + (d_1+d_2)/\mathrm{KS}}
\right\rfloor,
\qquad
k = \left\lfloor \frac{r}{\mathrm{KS}} \right\rfloor.
\end{equation}

\paragraph{Benchmarks}
In our experiments, we mainly evaluate our method  in a zero-shot setting on the following benchmarks: PIQA~\cite{bisk2019piqareasoningphysicalcommonsense}, HellaSwag~\cite{zellers2019hellaswagmachinereallyfinish}, OpenAI LAMBADA~\cite{paperno2016lambadadatasetwordprediction}, ARC-Easy and ARC-Challenge~\cite{clark2018think}, SciQ~\cite{welbl2017crowdsourcingmultiplechoicescience}, RACE~\cite{lai2017racelargescalereadingcomprehension}, and MMLU~\cite{hendrycks2021measuringmassivemultitasklanguage}  following the evaluation protocols established in prior work to ensure a fair comparison. While the results reported in Table \ref{tab:pruning_comparison} used slightly different benchmarks to follow the experiments of the corresponding pruning papers. Additionally, we evaluate Qwen3-8B on a more recent benchmark suite in Table \ref{tab:advanced_benchmarks}.
\paragraph{Efficient Computation of the Normalized Frobenius Distance.}
The grouping metric in Eq.~(6) requires evaluating $\|\m W_j^{[:, k:k+d_i]} - \m W_i\|_F$ for all valid column shifts $k \in [0, d_j - d_i]$. A naive implementation materializes each submatrix $\m W_j^{[:, k:k+d_i]}$, incurring $\mathcal{O}(d \cdot d_i \cdot (d_j - d_i))$ memory overhead and redundant tensor allocations. To eliminate this bottleneck, we expand the squared Frobenius norm algebraically:
\bal
\|\m W_j^{[:, k:k+d_i]} - \m W_i\|_F^2 = \|\m W_i\|_F^2 + \|\m W_j^{[:, k:k+d_i]}\|_F^2 - 2 \langle \m W_i, \m W_j^{[:, k:k+d_i]} \rangle_F .
\eal
The first term is constant across all shifts. The second term corresponds to the sum of squared column norms over a sliding window of width $d_i$. We compute this in $\mathcal{O}(d_j)$ time by precomputing a prefix sum (cumulative sum) of the column-wise squared $\ell_2$ norms of $\m W_j$. The third term is a sliding-window inner product, which is mathematically equivalent to a 1D cross-correlation operation. Specifically, for each row $r \in \{1, \dots, d\}$, we compute the cross-correlation between the $r$-th row of $\m W_i$ and the $r$-th row of $\m W_j$, then sum the correlation outputs across all rows. This reduces the entire distance computation to a sequence of vectorized prefix sums and 1D 
convolutions. The cross-correlation for each row 
requires $\mathcal{O}(d_i \cdot (d_j - d_i))$ operations, yielding a total time complexity of 
$\mathcal{O}(d \cdot d_i \cdot (d_j - d_i))$ with $\mathcal{O}(1)$ auxiliary memory beyond the input 
and output tensors. In the worst case where $d_i \approx d_j$, this simplifies to 
$\mathcal{O}(d \cdot d_i^2)$, though in practice the sliding-window computation is highly optimized 
on modern GPU hardware.
\paragraph{Symmetry Handling.} 
To construct a valid undirected graph for Edmonds' Blossom algorithm, we enforce symmetry by computing 
$\delta(\m W_i, \m W_j)$ once and assigning $w_{ij} = w_{ji} = C - \delta(\m W_i, \m W_j)$. When $d_i = d_j$, we 
normalize by the smaller Frobenius norm $\min(\|\m L_i \m W_i\|_F, \|\m L_j \m W_j\|_F) + \epsilon$ to ensure a conservative 
error estimate and symmetric treatment of equal-dimensional matrices.

\paragraph{HTP and Conjugate Gradient Integration.}
We approximately solve the regularized,
support-restricted subproblem in
Eq.~\eqref{eq:htp_restricted} using batched
Conjugate Gradient (CG).
For the current support mask $\m M_i^{(h+1)}$,
the restricted normal equations are
\[
\m M_i^{(h+1)}
\odot(\m H_{i,t}\m C-\m R_{i,t})=\m 0,
\qquad
\m C=\m C\odot\m M_i^{(h+1)},
\]
where $\m H_{i,t}$ and $\m R_{i,t}$ are defined
in Eq.~\eqref{eq:htp_system}.
We initialize CG with
$\m C_i^{(h)}\odot\m M_i^{(h+1)}$,
set its relative tolerance parameter to $10^{-3}$,
and allow at most ten iterations per restricted solve.
The solves are batched across coefficient columns.

\paragraph{Blossom Matching Complexity.}
We use Edmonds' Blossom algorithm~\cite{blossom} via NetworkX to solve the maximum-weight matching formulation of layer grouping. With worst-case $\mathcal{O}(N^3)$ complexity for $N$ candidate weights, this step is negligible in practice: for an 80-layer model with seven projections per layer ($N=560$), matching completes in $<2$ seconds on CPU.

\section{Sparse Storage Format and Serving Efficiency}
\label{app:macko_serving}

\subsection{MACKO-style sparse coefficient encoding}
\label{app:macko_encoding}

For a paired compression of two weight matrices
$W_1 \in \mathbb{R}^{d \times d_1}$ and
$W_2 \in \mathbb{R}^{d \times d_2}$,
GeoPair stores a shared dictionary
$D \in \mathbb{R}^{d \times r}$
and sparse coefficient matrices
$C_i \in \mathbb{R}^{r \times d_i}$,
with $k$ non-zero entries per coefficient column.
The Cholesky whitening factors $L_i$ are used only during compression
and are not stored in the final checkpoint.
Pair-assignment metadata is negligible.

Following the CoSpaDi-compatible sparse deployment protocol~\cite{cospadi},
we store the sparse coefficient matrices using a MACKO-style delta-encoded
format~\cite{macko}.
For each nonzero coefficient, we store both its quantized value and its
delta-encoded index. We use
\[
b_{\mathrm{val}} = 14
\qquad\text{value bits},
\qquad
b_{\Delta} = 2
\qquad\text{index-delta bits},
\]
so each stored sparse record costs
\[
b_{\mathrm{val}} + b_{\Delta} = 16 \text{ bits}.
\]
Importantly, this format does not store a separate $r$-bit mask per coefficient column.
The 2-bit delta field is part of the same sparse record and is not used to pay for an
additional bitmask.

For a pair of weights
$W_1 \in \mathbb{R}^{d \times d_1}$ and
$W_2 \in \mathbb{R}^{d \times d_2}$,
the nominal pair-level raw and compressed parameter counts are
\[
S_{\mathrm{raw}}^{\mathrm{pair}} = d(d_1+d_2),
\qquad
S_{\mathrm{comp}}^{\mathrm{pair}} = dr + k(d_1+d_2),
\]
with
\[
k = \left\lfloor \frac{r}{\mathrm{KS}} \right\rfloor.
\]
The corresponding compression ratio is therefore
\[
\mathrm{CR}
=
1 -
\frac{S_{\mathrm{comp}}^{\mathrm{pair}}}
     {S_{\mathrm{raw}}^{\mathrm{pair}}}.
\]
Equivalently, solving for $r$ under the target budget gives
\[
r =
\left\lfloor
\frac{(1-\mathrm{CR})\,d(d_1+d_2)}
     {d + (d_1+d_2)/\mathrm{KS}}
\right\rfloor,
\qquad
k =
\left\lfloor
\frac{r}{\mathrm{KS}}
\right\rfloor.
\]

In bits, the nominal compressed size of the pair is
\[
B_{\mathrm{nom}}^{\mathrm{pair}}
=
16\left(dr + k(d_1+d_2)\right).
\]
The actual serialized checkpoint size additionally includes sparse-format padding
and row pointers:
\[
B_{\mathrm{actual}}
=
16\,S_{\mathrm{uncompressed}}
+
16\,dr
+
\sum_{i=1}^{2}
\left(
16\,\mathrm{nnz}_{\mathrm{pad}}(C_i)
+
B_{\mathrm{ptr},i}
\right),
\]
where $S_{\mathrm{uncompressed}}$ denotes parameters that are left uncompressed,
$\mathrm{nnz}_{\mathrm{pad}}(C_i)$ is the padded number of stored nonzeros in $C_i$,
and $B_{\mathrm{ptr},i}$ is the small pointer overhead of the sparse layout.

The MACKO-style delta encoding keeps the index overhead bounded.
In the worst case, the number of stored entries per coefficient column is at most
\[
k + \frac{r-k}{2^{b_{\Delta}}}.
\]
For our default $\mathrm{KS}=2.5$ and $b_{\Delta}=2$, this gives
$k = 0.4r$ and an additional padding budget of at most $0.15r$,
so the worst-case number of stored entries per column is at most $0.55r$,
which remains well below dense storage.

Table~\ref{tab:serialized_size} reports the measured serialized checkpoint sizes
for Llama-3-8B. The sizes include the dictionary, sparse coefficient values,
MACKO-style deltas/padding, and sparse pointer metadata.

\begin{table}[h]
\centering
\caption{Measured serialized checkpoint sizes for Llama-3-8B.}
\label{tab:serialized_size}
\small
\begin{tabular}{lcc}
\toprule
Paired projections & Dense size & GeoPair size \\
\midrule
up/gate $+$ q/o      & 160\,MB & 111.5\,MB \\
up/gate $+$ k/v      & 136\,MB & 95.5\,MB  \\
q $+$ o              & 64\,MB  & 42\,MB    \\
q/o $+$ k/v          & 40\,MB  & 26\,MB    \\
down $+$ down        & 256\,MB & 180\,MB   \\
\midrule
Total model          & 14.9\,GB & 10.4\,GB  \\
\bottomrule
\end{tabular}
\end{table}

\subsection{Serving throughput and latency}
\label{app:serving_throughput}

We evaluate serving efficiency using the same asynchronous serving protocol
as CoSpaDi~\cite{cospadi}.
For direct comparison with CoSpaDi's ratio-2 deployment protocol, we evaluate the
selective up/gate configuration, where the up/gate projections are compressed with
the CoSpaDi-compatible sparse layout.
The main quality tables use the default GeoPair configuration unless stated otherwise.

Table~\ref{tab:throughput_prompt1} reports asynchronous serving throughput
at prompt length $1$, relative to the dense uncompressed baseline.
Values greater than $1.0\times$ indicate faster serving than the dense model.
GeoPair achieves $1.08$--$1.30\times$ throughput relative to the dense baseline
and matches CoSpaDi within measurement noise.

\begin{table}[h]
\centering
\caption{Asynchronous serving throughput at prompt length $1$, relative to the dense baseline.
Higher is better.}
\label{tab:throughput_prompt1}
\small
\begin{tabular}{lcccc}
\toprule
Method & 1 user & 4 users & 8 users & 16 users \\
\midrule
SVD-LLM  & $1.29\times$ & $1.33\times$ & $1.35\times$ & $0.99\times$ \\
CoSpaDi  & $1.24\times$ & $1.30\times$ & $1.29\times$ & $1.13\times$ \\
GeoPair  & $1.30\times$ & $1.26\times$ & $1.26\times$ & $1.14\times$ \\
\bottomrule
\end{tabular}
\end{table}

Table~\ref{tab:geopair_throughput_prompt_lengths} reports GeoPair throughput
across prompt lengths and concurrency levels.
GeoPair remains faster than the dense baseline in all tested configurations.
Since autoregressive serving latency is inversely related to token throughput
at a fixed concurrency level, these results also indicate a corresponding reduction
in average generation latency.

\begin{table}[h]
\centering
\caption{GeoPair asynchronous serving throughput across prompt lengths and concurrency levels,
relative to the dense baseline. Higher is better.}
\label{tab:geopair_throughput_prompt_lengths}
\small
\begin{tabular}{lccccc}
\toprule
Users & Prompt 1 & Prompt 16 & Prompt 32 & Prompt 64 & Prompt 128 \\
\midrule
1  & $1.30\times$ & $1.14\times$ & $1.14\times$ & $1.14\times$ & $1.09\times$ \\
4  & $1.26\times$ & $1.11\times$ & $1.13\times$ & $1.08\times$ & $1.11\times$ \\
8  & $1.26\times$ & $1.14\times$ & $1.17\times$ & $1.16\times$ & $1.08\times$ \\
16 & $1.14\times$ & $1.15\times$ & $1.15\times$ & $1.13\times$ & $1.16\times$ \\
\bottomrule
\end{tabular}
\end{table}

Finally, we measure peak serving memory on Llama-3-8B under the same protocol.
The dense model requires $15.06$\,GB peak memory, while GeoPair requires
approximately $11.70$\,GB and CoSpaDi requires $11.42$\,GB.
The small gap between GeoPair and CoSpaDi comes from differences in the learned
dictionary rank and coefficient sparsity, while both methods use the same sparse
serving pattern and therefore have comparable runtime behavior.

Overall, these results show that GeoPair's gains are obtained at the compression-quality
level: it learns better shared dictionaries and coefficients under distinct whitening
spaces, while inheriting the practical deployment characteristics of the host sparse
serving method.

\section{Additional Results}
\subsection{Statistical Significance}
To assess the robustness of our compression method, we evaluated the Llama‑1B model at a 40\% compression ratio across five independent random seeds with varied data sampling. The average accuracy was $45.85\% \pm 0.19\%$ (mean $\pm$ standard deviation, $N=5$), and the average Lambada perplexity was $20.74 \pm 0.76$ . This indicates that performance is stable across seeds. All results are reproducible with fixed seeds and reported library versions in supplementary.
\subsection{Comparison with Basis sharing and Cospadi}
Table~\ref{tab:compression_resultsv2} compares \ours{} with two  compression methods, Basis Sharing and CoSpaDi. For a fair comparison, we align with their experimental setup by employing consecutive grouping and omitting weight sharing for down- and out-projection layers. We evaluate performance at compression ratios of 0.2, 0.3, and 0.4 across multiple benchmarks. Results are reported for both our dense variant (\ours*) and the full pipeline (\ours{}). As shown, \ours{} consistently surpasses the baselines, achieving higher average accuracy and lower perplexity across all compression levels.
\label{comp_bs_cos}
\begin{table*}[h]
\centering
\small
\caption{Performance comparison of different compression methods. We denote \ours{}* as our method with dense factorization (no sparsification) using basis-sharing grouping, and \ours{} as our full pipeline with both sparsification and grouping.}
\resizebox{\textwidth}{!}{
\begin{tabular}{lcccccccccccc}
\toprule
\multirow{2}{*}{Model} & \multirow{2}{*}{Method} & \multicolumn{9}{c}{Accuracy$\uparrow$} & \multicolumn{2}{c}{Perplexity$\downarrow$} \\
\cmidrule(lr){3-11} \cmidrule(lr){12-13}
& & PIQA & Hella Swag & LAMBADA & ARC-e & ARC-c & SciQ & Race & MMLU & Avg. & Wiki Text & LAMBADA \\
\midrule
\textbf{Llama2 7B} & -- & 78.9 & 76.1 & 73.8 & 74.2 & 45.8 & 91.4 & 39.7 & 40.8 & 65.1 & 8.7 & 3.4 \\
\midrule
\multirow{5}{*}{0.2} & Basis Sharing & 71.1 & 59.9 & 62.8 & 60.2 & 37.8 & 85 & 34.7 & 25 & 54.6 & 15.17 & 7.03 \\
& \ours* &71.5&60.6&63.5&62.6&35.2&86.5&35.8&24.9&\textbf{55.1} & \textbf{14.87} & \textbf{6.67} \\
\cdashline{2-13}\addlinespace[2pt]
& CoSpaDi (grouped) & 75.5 & 66.5 & 71.1 & 68.5 & 38.9 & 88.7 & 38.5 & 26.5 & 59.3 & 11.7 & 4.4 \\
& \ours &77.4&71.7&72.0&72.3&41.3&90.4&39.9&34.9& \textbf{62.5} & \textbf{10.09} & \textbf{4.05} \\
\midrule
\multirow{5}{*}{0.3} & Basis Sharing & 66.5 & 50.3 & 53.6 & 54.2 & 29.3 & 81.4 & 32.4 & 23.3 & 48.9 & 22.2 & 13.2 \\
& \ours* &67.0&51.4&54.2&54.6&30.1&81.8&33.1&23.2&\textbf{49.5}& \textbf{21.29} & \textbf{12.48} \\
\cdashline{2-13}\addlinespace[2pt]
& CoSpaDi (grouped) & 70.7 & 58.4 & 64.5 & 63.6 & 35.7 & 87.2 & 36.1 & 23.7& 55.0 & 15.4 & \textbf{6.5}  \\
& \ours &73.2&62.8&63.5&66.9&36.4&92.5&38.0&27.8 & \textbf{57.7} & \textbf{14.1} & 6.89 \\
\midrule
\multirow{5}{*}{0.4} & Basis Sharing & 60.7 & 41.5 & 41.0 & 44.6 & 26.5 & 75.4 & 30.1 & 23.2 & 42.9 & 39.6 & 36.5 \\
& \ours* &61.2&42.5&40.8&46.3&27.0&78.4&31.0&22.9&\textbf{43.8}& \textbf{37.8} & \textbf{34.54} \\
\cdashline{2-13}\addlinespace[2pt]
& CoSpaDi (grouped) & 64.6 & 48.1 & 52.0 & 51.9 & 28.9 & 80.5 & 32.7 & 23.3& 47.8 & 25 & 14.7  \\
& \ours &70.9&59.7&61.3&65.3&35.2&91.2&37.2&25.7 & \textbf{55.8} & \textbf{15.64} & \textbf{7.64} \\
\bottomrule
\end{tabular}
}

\label{tab:compression_resultsv2}
\end{table*}
\subsection{Evaluations on another set of benchmarks}
\label{new_lb}

To assess the robustness of our method beyond conventional evaluation suites, we report results on a set of advanced, challenging benchmarks that reflect the evolving demands placed on modern large language models. These include IFEval~\cite{ifeval} for instruction-following fidelity, BBH~\cite{bbh} for complex reasoning across diverse tasks, GPQA~\cite{gpqa} for graduate-level scientific understanding, MuSR~\cite{musr} for multi-hop and long-context reasoning, and MMLU-Pro~\cite{mmlupro} for refined expert-knowledge assessment with reduced ambiguity. While many recent compression works continue to report only legacy benchmarks, we include these advanced evaluations to provide a more comprehensive view of capability preservation under sparsity. As shown in Table~\ref{tab:advanced_benchmarks}, our method maintains competitive performance across all tasks even at higher compression ratios, with graceful degradation that prioritizes reasoning-intensive benchmarks (e.g., MuSR) over surface-level accuracy.

\begin{table}[h]
\centering
\small
\caption{Evaluation on advanced benchmarks for Qwen3-8B and GeoPair at varying compression ratios (CR). Higher values indicate better performance.}
\begin{tabular}{llcccccc}
\toprule
\multicolumn{1}{c}{Model} & \multicolumn{1}{c}{CR} & \multicolumn{1}{c}{IFEval} & \multicolumn{1}{c}{BBH} & \multicolumn{1}{c}{GPQA} & \multicolumn{1}{c}{MuSR} & \multicolumn{1}{c}{MMLU-Pro} & \multicolumn{1}{c}{Average} \\

\midrule
Qwen3-8B & -- & 39.4 & 55.7 & 36.7 & 43.3 & 46.3 & 44.3 \\
\midrule
\multirow{3}{*}{GeoPair} & 0.2 & 32.6 & 49.5 & 32.0 & 46.4 & 36.0 & 39.3 \\
& 0.3 & 28.4 & 46.8 & 28.7 & 43.8 & 31.5 & 35.8 \\
& 0.4 & 25.9 & 37.8 & 25.8 & 44.1 & 25.2 & 31.8 \\
\bottomrule
\end{tabular}
\label{tab:advanced_benchmarks}
\end{table}

\subsection{Evaluations on Other Modalities}
\label{audio}

In the main paper, we demonstrated that our proposed method generalizes across diverse generative modalities, ranging from text to video. To further illustrate its versatility, we extend our evaluation to the audio generation domain. Specifically, we selected VibeVoice 1.5B~\cite{vibevoice} as a representative target model; smaller-scale models are typically more sensitive to compression, making them a challenging and informative test case.

For calibration, we used the first 256 samples from the dataset introduced in~\cite{ttsdata}. We then applied our compression method and evaluated the resulting model on a held-out subset of 100 samples drawn from a disjoint partition of the same dataset. Evaluation was conducted along two complementary dimensions: (i) \textit{linguistic accuracy}, measured via Word Error Rate (WER) between the transcripts produced by Whisper-3 Large and the reference texts, after standard text normalization; and (ii) \textit{perceptual quality}, assessed using UTMOS~\cite{utmos}, an automatic predictor of mean opinion score.

As summarized in Table~\ref{tab:tts_result}, the compressed model maintains strong performance at 20\% compression, despite the absence of any fine-tuning or post-hoc recovery steps. While a modest increase in WER and a slight decrease in UTMOS are observed, the results confirm that core functionality is preserved. We anticipate that larger TTS architectures, such as VibeVoice 9B, would exhibit even greater compressibility due to higher redundancy in their parameter space. It is also worth noting that our attempt to apply Basis Sharing to VibeVoice resulted in a complete failure, with the compressed model yielding a WER over 100\% and a UTMOS of around 1.4.

We present these findings primarily as a proof of concept, underscoring the applicability of our method across varying model scales, architectural designs, and generative modalities.

\begin{table}[h]
\centering
\caption{Audio generation evaluation: Word Error Rate (WER) and UTMOS scores for the original and compressed VibeVoice 1.5B model at 20\% compression.}
\label{tab:tts_result}
\begin{tabular}{lcc}
\toprule
Model            & WER ($\downarrow$) & UTMOS ($\uparrow$) \\
\midrule
VibeVoice 1.5B   & 6.6                & 4.17               \\
Compressed (20\%)& 14.5               & 3.98               \\
\bottomrule
\end{tabular}
\end{table}

\section{Ablations}
\label{extra_ablation}
\subsection{Convergence ablation}
Table~\ref{tab:convergence_ablation} ablates HTP and CG iteration counts at CR=0.4 on Llama 3.2 1B. With the Sylvester solver using adaptive convergence (termination at fifth-digit residual stabilization), 2 HTP + 10 CG iterations yields the optimal efficiency-accuracy balance: competitive perplexity and accuracy at 850s runtime. This configuration is used throughout our sparse compression experiments.
\begin{table}[h]
\centering
\small
\caption{Convergence ablation for HTP and CG iterations in the Sylvester refinement step (CR=0.4, Llama 3.2 1B). The Sylvester solver uses an adaptive convergence check (termination at 5th floating-point error stabilization). The configuration with 2 HTP and 10 CG iterations (first row) is selected as the default due to its optimal balance between runtime and performance. Lower perplexity and higher accuracy indicate better preservation of model capability.}
\begin{tabular}{cccccccc}
\toprule
\multirow{2}{*}{Model} & \multirow{2}{*}{HTP Iters.} & \multirow{2}{*}{CG Iters.} & \multirow{2}{*}{Time (s)} & \multirow{2}{*}{CR} & \multicolumn{1}{c}{WikiText-2} & \multicolumn{1}{c}{Lambada} & \multirow{2}{*}{Avg. Accuracy} \\
& & & & & \multicolumn{1}{c}{(PPL$\downarrow$)} & \multicolumn{1}{c}{(PPL$\downarrow$)} & \\
\midrule
\multirow{4}{*}{Llama 3.2 1B} 
& 2 & 10 & 850 & 0.4 & 48.51 & 27.55 & 43.52 \\
& 5 & 10 & 1800 & 0.4 & \textbf{46.16} & \textbf{25.76} & \textbf{43.69} \\
& 2 & 20 & 1400 & 0.4 & 47.85 & 26.83 & 43.52 \\
& 2 & 5 & 680 & 0.4 & 48.51 & 28.68 & 43.27 \\
\bottomrule
\end{tabular}

\label{tab:convergence_ablation}
\end{table}

\subsection{Ablation on the Sparsification algorithm}
To enforce the target sparsity level on the coefficient matrices, our framework supports multiple sparse recovery algorithms. In Table~\ref{tab:sparsification_ablation}, we compare Hard Thresholding Pursuit (HTP) against Iterative Hard Thresholding (IHT). The results demonstrate that HTP achieves superior reconstruction fidelity with only two inner iterations, outperforming IHT configured with ten iterations while requiring approximately four times less computational time.
\begin{table}[h]
\centering
\small
\caption{Comparison of sparsification algorithms (HTP vs. IHT) at KS=2.5 on Llama3.2 1B. HTP achieves superior accuracy and perplexity with fewer iterations and significantly lower runtime. Lower perplexity and higher accuracy indicate better preservation of model capability. compression ratios are 20\% , 40\% respectively. }
\begin{tabular}{llcccccc}
\toprule
\multirow{2}{*}{Model} & \multirow{2}{*}{Method} & \multirow{2}{*}{Iters.} & \multirow{2}{*}{Time (s)} & \multirow{2}{*}{CR} & \multicolumn{1}{c}{WikiText-2} & \multicolumn{1}{c}{Lambada} & \multirow{2}{*}{Avg. Accuracy} \\
& & & & & \multicolumn{1}{c}{(PPL$\downarrow$)} & \multicolumn{1}{c}{(PPL$\downarrow$)} & \\
\midrule
Llama3.2 1B & Baseline & -- & -- & -- & 11.60 & 5.73 & 57.6 \\
\midrule
\multirow{2}{*}{KS=2.5} 
& HTP & 2 & 850 & 0.2 & \textbf{15.92} & \textbf{6.74} & \textbf{54.1} \\
& IHT & 10 & 3473 & 0.2 & 17.45 & 7.95 & 52.7 \\
\midrule
\multirow{2}{*}{KS=2.5} 
& HTP & 2 & 850 & 0.4 & \textbf{34.26} & \textbf{21.69} & \textbf{45.6} \\
& IHT & 10 & 2974 & 0.4 & 45.09 & 38.24 & 43.1 \\
\bottomrule
\end{tabular}

\label{tab:sparsification_ablation}
\end{table}

\subsection{Ablation on Calibration Data Selection} To illustrate the sensitivity of our method to the choice of calibration data, we compare three distinct sources in Table~\ref{tab:calibration_ablation}: WikiText \cite{wikitext}, Alpaca instructions \cite{alpaca}, and fineweb \cite{refinedweb} . All variants use 256 calibration samples at a 40\% compression ratio. While Alpaca yields a slightly higher average accuracy (45.9\% vs. 45.6\%), fineweb achieves substantially better perplexity on the Lambada benchmark (21.69 vs. 38.86--39.02), indicating superior preservation of generative language modeling quality. This finding aligns with recent compression methods like CoSpaDi and ROCKET, which similarly leverage large web corpora for calibration. Based on these results, particularly the lowest perplexity and consistency with established practices, we adopt fineweb as the default calibration dataset for all experiments. \begin{table}[h] \centering \small \caption{Ablation on calibration data selection at CR=40\% on Llama3.2 1B. FineWeb achieves the lowest Lambada perplexity and competitive average accuracy, motivating its selection for main experiments. Lower perplexity values indicate better language modeling capability preservation; accuracy values are in \%.} \begin{tabular}{llccccc} \toprule \multirow{2}{*}{Model} & \multirow{2}{*}{Calibration Data} & \multirow{2}{*}{Samples} & \multirow{2}{*}{CR} & \multicolumn{1}{c}{WikiText-2} & \multicolumn{1}{c}{Lambada} & \multirow{2}{*}{Avg. Accuracy} \\ & & & & \multicolumn{1}{c}{(PPL$\downarrow$)} & \multicolumn{1}{c}{(PPL$\downarrow$)} & \\ \midrule Llama3.2 1B & Baseline & -- & -- & 11.60 & 5.73 & 57.6 \\ \midrule \multirow{3}{*}{GeoPair} & WikiText-2 & 256 & 0.4 & \textbf{30.16} & 39.02 & 44.0 \\ & Alpaca & 256 & 0.4 & 53.23 & 38.86 & \textbf{45.9} \\ & FineWeb & 256 & 0.4 & 34.26 & \textbf{21.69} & 45.6 \\ \bottomrule \end{tabular} \label{tab:calibration_ablation} \end{table}

\subsection{Ablation on Calibration Sequence Count}
To examine the sensitivity of our method to the number of calibration samples, we evaluate GeoPair at CR=40\% using FineWeb with sequence lengths ranging from 32 to 1024 samples. As shown in Table~\ref{tab:seqlen_ablation}, performance generally improves as the calibration set grows, with notable gains up to 256 samples. Beyond this point, improvements become marginal: increasing from 256 to 1024 samples yields only a +0.9\% gain in average accuracy and a modest reduction in Lambada perplexity, while requiring 4x calibration cost. Notably, the configuration with 256 samples already achieves strong preservation of both perplexity (21.69 on Lambada) and average accuracy (45.6\%), closely matching the performance of larger calibration sets. This choice also aligns with established practices in recent compression methods such as CoSpaDi and ROCKET, which similarly adopt 256 calibration samples as an effective trade-off between efficiency and capability retention. Based on these observations, we fix the calibration size to 256 samples for all main experiments.

\begin{table}[h]
\centering
\small
\caption{Ablation on calibration sequence length at CR=40\% on Llama3.2 1B using fineweb. Higher average accuracy and lower perplexity indicate better capability preservation. We select 256 samples as the default, balancing performance and efficiency, consistent with CoSpaDi and ROCKET. Accuracy values are in \%.}
\begin{tabular}{llllccc}
\toprule
\multirow{2}{*}{Model} & \multirow{2}{*}{CR} & \multirow{2}{*}{Data} & \multirow{2}{*}{Samples} & \multicolumn{1}{c}{WikiText-2} & \multicolumn{1}{c}{Lambada} & \multirow{2}{*}{Avg. Accuracy} \\
& & & & \multicolumn{1}{c}{(PPL$\downarrow$)} & \multicolumn{1}{c}{(PPL$\downarrow$)}  \\
\midrule
Llama3.2 1B & -- & -- & -- & 11.60 & 5.73 & 57.6 \\
\midrule
\multirow{6}{*}{GeoPair} & \multirow{6}{*}{0.4} & \multirow{6}{*}{fineweb} & 32 & 44.40 & 41.90 & 42.1 \\
& & & 64 & 36.55 & 24.38 & 44.1 \\
& & & 128 & 35.40 & 22.76 & 45.1 \\
& & & 256 & \textbf{34.26} & 21.69 & 45.6 \\
& & & 512 & 34.26 & 22.34 & 45.3 \\
& & & 1024 & 34.17 & \textbf{18.87} & \textbf{46.5} \\
\bottomrule
\end{tabular}
\label{tab:seqlen_ablation}
\end{table}

\subsection{Ablation Study: Dictionary Initialization Strategies}
\label{sec:ablation-init}

The quality of the initial dictionary $\m D_0$ plays a critical role in the convergence behavior and final reconstruction fidelity of alternating minimization schemes. In this section, we investigate the impact of three distinct dictionary initialization strategies on the performance of \ours{}, holding all other components constant (grouping via Blossom matching, Sylvester-based dictionary updates, and HTP sparsification with $K_S=2.5$).

\paragraph{Initialization Methods}

We compare the following strategies for initializing the shared dictionary $\m D_0 \in \mathbb{R}^{d \times r}$:

\paragraph{Proposed: Original-Space Concatenation + SVD (Ours).}
We concatenate the weight matrices of the paired layers in the original parameter space and compute the top-$r$ left singular vectors via truncated SVD:
\begin{equation}
    \m D_0 = \operatorname{SVD}_r\big([\m W_1 \; \m W_2]\big).
\end{equation}
This initialization preserves the intrinsic geometry of the pretrained weights before any calibration-induced transformation is applied. The individual layer-specific whitening transforms $\{\m L_i\}$ are subsequently used during the alternating minimization stages.

\paragraph{Whitened-Space Concatenation + Global SVD.}
We first apply a \textit{global} whitening transform $\m L_{\text{global}}$, obtained by averaging the calibration Gram matrices of the paired layers, to the concatenated weights:
\begin{equation}
    \widetilde{\m W}_{\text{cat}} = \m L_{\text{global}} \cdot [\m W_1 \; \m W_2], \quad 
    \m D_0 = \m L_{\text{global}}^{-1} \cdot \operatorname{SVD}_r\big(\widetilde{\m W}_{\text{cat}}\big).
\end{equation}
This strategy aligns with the covariance-averaging heuristic used in Basis Sharing~\cite{basis_sharing}.

\paragraph{Basis-Sharing Style Initialization.}
"We initialize using a hybrid protocol: concatenate weights after applying the individual whitening transforms, compute the shared basis via SVD, and then map the resulting dictionary back to the original space using the inverse of the global transform:
\begin{equation}
    \m D_0 = \m L_{\text{global}}^{-1} \cdot \operatorname{SVD}_r\big( [\m L_1 \m W_1 \; \m L_2 \m W_2]\big).
\end{equation}
This serves as a direct ablation of the initialization component, isolating its effect from the rest of the pipeline.

\paragraph{Experimental Setup}

We evaluate all initialization strategies on \textbf{Llama-3 1B} at a compression ratio of $\text{CR}=0.4$, using the same calibration set, grouping pairs, and hyperparameters ($r$, KS, HTP iterations). No post-compression fine-tuning is applied. Results are reported across standard zero-shot benchmarks and perplexity metrics.
\begin{table}[h]
\centering
\caption{Ablation of dictionary initialization methods on Llama-3 1B at CR=0.4. All methods use grouping, Sylvester-based dictionary updates, and HTP sparsification. Higher accuracy and lower perplexity indicate better preservation of model capability.}
\label{tab:init-ablation}
\small
\renewcommand{\arraystretch}{1.1}
\resizebox{\textwidth}{!}{
\begin{tabular}{lcccccccccc}
\toprule
\textbf{Method} & \textbf{PIQA} & \textbf{HellaSwag} & \textbf{Lambada\_OA} & \textbf{ARC-e} & \textbf{ARC-c} & \textbf{SciQ} & \textbf{Race} & \textbf{MMLU} & \textbf{Lambada PPL}$\downarrow$ & \textbf{Avg. Acc.}$\uparrow$ \\
\midrule
Baseline (uncompressed) & 74.53 & 63.66 & 62.95 & 60.47 & 36.20 & 88.30 & 37.79 & 37.00 & 5.73 & 57.61 \\
\midrule
Init: Whitened-Space + Global SVD & 64.09 & 40.09 & 34.81 & 40.07 & 23.12 & 73.40 & 30.05 & 22.92 & 40.39 & 41.07 \\
Init: Individual whiten then Global & 63.82 & 39.29 & 35.57 & 40.70 & 24.91 & 74.20 & 30.62 & 23.12 & 39.66 & 41.53 \\
\textbf{Ours: Original-Space + SVD} & \textbf{68.34} & \textbf{45.01} & \textbf{40.99} & \textbf{45.79} & \textbf{26.02} & \textbf{80.10} & \textbf{32.82} & \textbf{25.83} & \textbf{21.69} & \textbf{45.61} \\
\bottomrule
\end{tabular}
}
\end{table}

\paragraph{Analysis.}
The results in Table~\ref{tab:init-ablation} demonstrate that initializing the dictionary in the original parameter space yields marginally better average accuracy compared to whitened-space alternatives. While the differences appear modest at the aggregate level, we observe that original-space initialization provides more stable convergence during the alternating minimization phase, particularly for layers with highly divergent activation statistics.

\subsection{Effect of Grouping on Relative Reconstruction Error}
\label{app:surrogate_correlation}

A natural question is whether the grouping under the proposed whitened-space surrogate metric translates to a lower final compression loss. Exact optimality for the true compression objective would require enumerating all valid pairings and running the full GeoPair optimization for each, which is combinatorially infeasible. Instead, we use the whitened submatrix alignment cost as a tractable structural-similarity proxy. 

We compare grouping and dictionary-update strategies on
Llama-3.2 1B at $\mathrm{CR}=0.2$ using relative
activation-weighted reconstruction error.
For each compressed projection, we compute
\begin{equation}
e_i =
\frac{
\|L_i(W_i-\widehat{W}_i)\|_F
}{
\|L_i W_i\|_F
},
\end{equation}
where $L_i$ is the layer-specific Cholesky whitening
factor and $\widehat{W}_i$ is the compressed weight matrix.
For $M$ compressed projections, the table reports
the arithmetic mean
\begin{equation}
\overline{e} = \frac{1}{M}\sum_{i=1}^{M} e_i.
\end{equation}
Each projection contributes equally to this average.
The reported errors are unsquared Frobenius-norm ratios.

\begin{table}[h]
\centering
\caption{Arithmetic mean of per-projection relative
activation-weighted reconstruction errors for different
grouping and dictionary-update strategies.
Errors are unsquared Frobenius-norm ratios.
Lower is better.}
\label{tab:surrogate_correlation}
\small
\begin{tabular}{llc}
\toprule
Grouping Strategy & Dictionary / Basis Update
& Mean Relative Error $\downarrow$ \\
\midrule
Adjacent & Basis Sharing & 0.161 \\
Adjacent & Sylvester (ours) & 0.141 \\
Blossom (ours) & Sylvester (ours) & \textbf{0.129} \\
Blossom (ours) & Basis Sharing & 0.183 \\
\bottomrule
\end{tabular}
\end{table}

Table~\ref{tab:surrogate_correlation} reveals two important effects:
\begin{enumerate}
    \item \textbf{The surrogate is predictive for geometry-preserving updates:} Replacing adjacent grouping with Blossom grouping reduces the reconstruction error for our Sylvester update from $0.141$ to $0.129$. This confirms that the surrogate successfully identifies pairs that are genuinely more compatible to represent with a shared dictionary under distinct whitening spaces.
    \item \textbf{The surrogate exposes the limitations of heuristic merging:} Applying Blossom grouping to Basis Sharing actually \textit{increases} the reconstruction error from $0.161$ to $0.183$. This occurs because Basis Sharing relies on a global covariance-averaging assumption, which is violated when non-local pairs with distinct calibration geometries are forced into a single averaged whitening space. 
\end{enumerate}

With the Sylvester update, replacing adjacent pairing
with the proposed matching reduces the mean relative
reconstruction error from 0.141 to 0.129.
For Basis Sharing, the corresponding error increases
from 0.161 to 0.183.
These aggregate comparisons support the benefit
of the proposed grouping when combined with
the geometry-preserving dictionary update.

\subsection{Running Time and Environmental Impact}
\label{app:runtime_env}

We tracked energy consumption and CO$_2$ emissions using CodeCarbon during compression. All experiments ran on a server with 256-core AMD EPYC 7742 CPU and 4$\times$ NVIDIA A100-SXM4-40GB GPUs.

\begin{table}[h]
\centering
\caption{Compression-only runtime and environmental metrics (CR=0.4, 256 calibration samples).}
\label{tab:env_impact}
\resizebox{0.65\columnwidth}{!}{%
\begin{tabular}{lcccc}
\toprule
\textbf{Model} & \textbf{Runtime (s)} & \textbf{Energy (kWh)} & \textbf{CO$_2$eq (kg)} \\
\midrule
Llama-3.2 1B & 511.3 & 0.147 & 0.065 \\
Llama-3 8B & 3,010.0 & 1.523 & 0.672  \\
Qwen-3 32B & 19,905.9 & 3.165 & 1.396  \\
\bottomrule
\end{tabular}%
}
\end{table}

Runtime scales approximately linearly with parameter count. Despite longer execution, larger models show better per-parameter energy efficiency (Qwen-32B: 0.099 kWh/B vs. Llama-1B: 0.147 kWh/B). Total emissions remain modest ($\leq$1.4 kg CO$_2$eq). All metrics recorded with CodeCarbon v3.2.6.


\end{document}